\PassOptionsToPackage{protrusion=false}{microtype}
\documentclass[11pt]{bluue}

\usepackage{amsmath,amsfonts,bm}

\def\eqref#1{equation~\ref{#1}}

\def\1{\bm{1}}

\DeclareMathAlphabet{\mathsfit}{\encodingdefault}{\sfdefault}{m}{sl}
\SetMathAlphabet{\mathsfit}{bold}{\encodingdefault}{\sfdefault}{bx}{n}

\usepackage{hyperref}
\usepackage{url}
\usepackage{graphicx}
\usepackage{booktabs}
\usepackage{multirow}
\usepackage{amsmath,amssymb}
\usepackage{xspace}
\usepackage{makecell}
\usepackage{wrapfig}
\usepackage{textcomp,fontawesome5,listings,upquote}
\usepackage{tabularx}
\tcbuselibrary{skins,breakable}
\definecolor{envschedblue}{RGB}{220,232,245}
\definecolor{deepgreen}{RGB}{0, 70, 0}
\definecolor{deepred}{RGB}{182, 32, 22}
\definecolor{jsongreen}{RGB}{0, 153, 0}
\definecolor{lightblue}{RGB}{230, 246, 250}
\newcommand{\envsched}{\textsc{EnvSched}\xspace}
\newsavebox{\arxivdependencybox}
\AtBeginDocument{%
\fancyhead[C]{\makebox[\headwidth][s]{%
  \includegraphics[height=\instlogoheight]{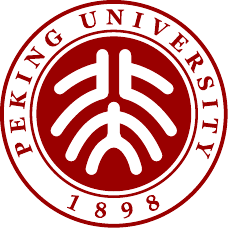}%
  \hfill
  \includegraphics[height=\instlogoheight]{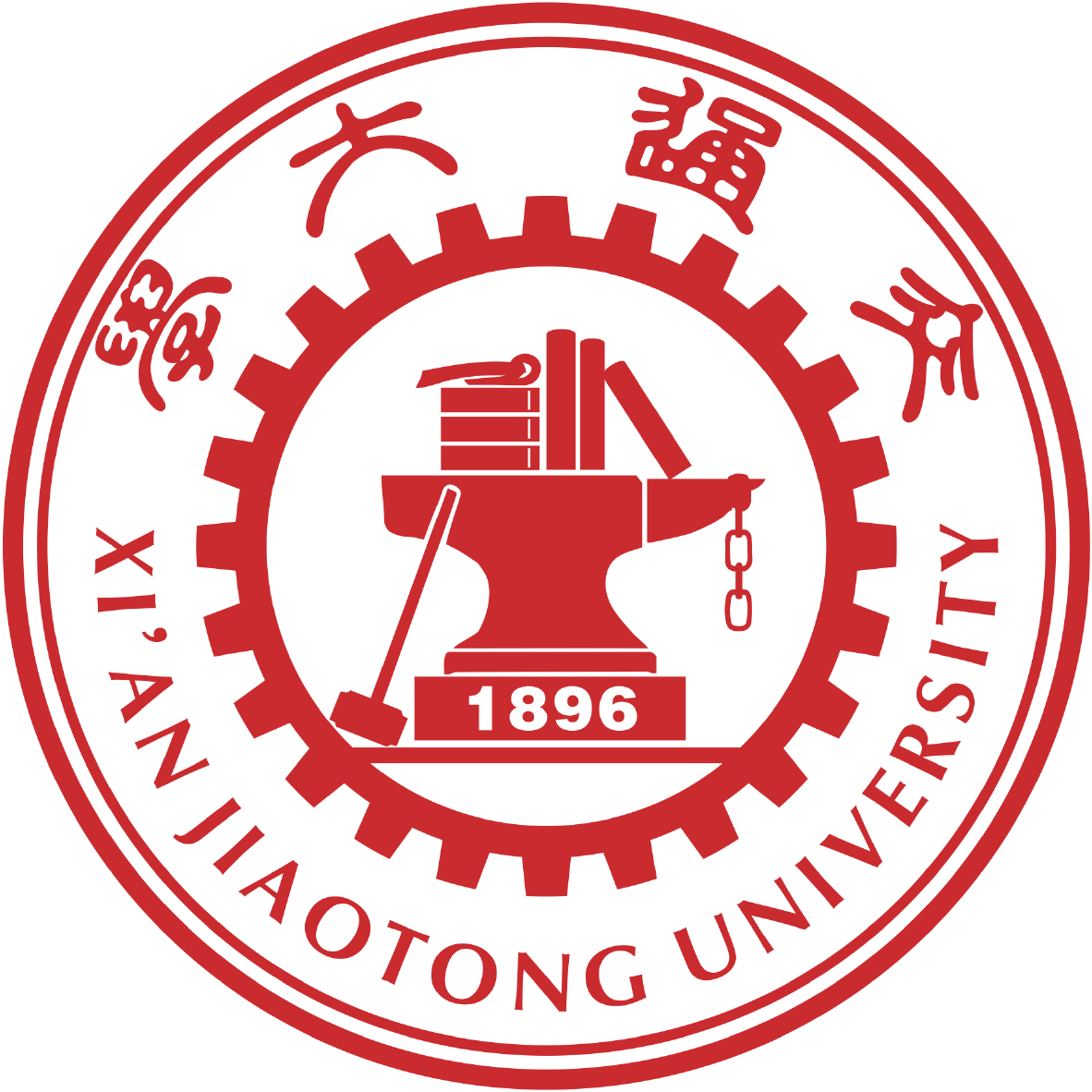}%
  \hfill
  \includegraphics[height=\instlogoheight]{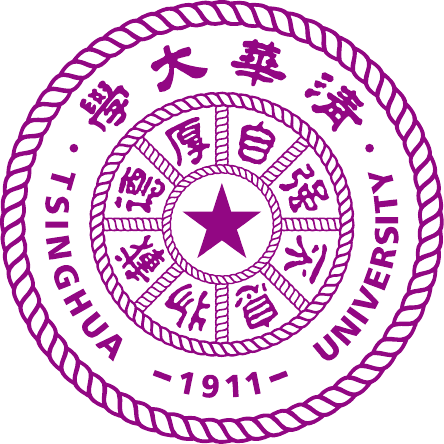}%
  \hfill
  \includegraphics[height=\instlogoheight]{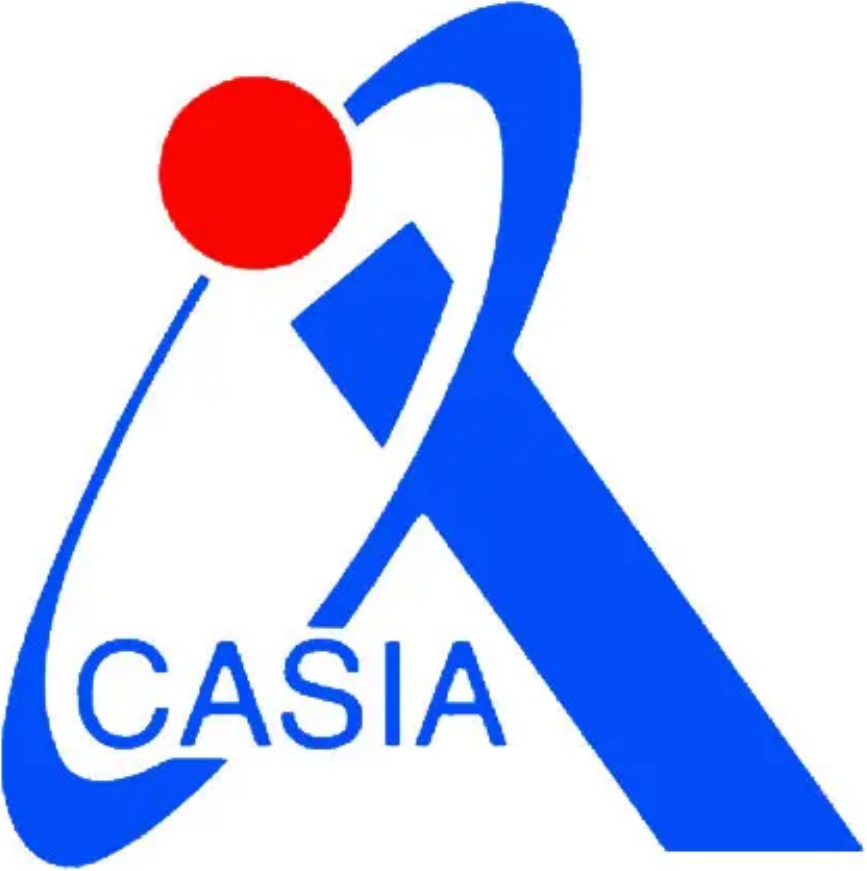}%
  \hfill
  \includegraphics[height=\instlogoheight]{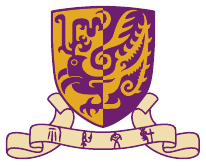}%
}}
}

\title{EnvSched: Environment-Aware Scheduling for Rollout-Efficient Agent Reinforcement Learning}

\author[1,*]{Sihan Yang}  
\author[2,*]{Weichen Bai}
\author[1,*,\dagger]{Ruichuan An}
\author[3]{Yiming Cheng}
\authorbreak
\author[3]{Zhiyuan Feng}
\author[1]{Tianyi Jiang}
\author[1]{Renye Yan}
\author[4]{Ning Ding}
\authorbreak
\author[5,\dagger]{Yifan Zhang}
\author[6]{Renrui Zhang}
\author[4,\ddagger]{Yehui Tang}
\author[1,\ddagger]{Wentao Zhang}

\affiliation[1]{PKU}  
\affiliation[2]{XJTU}  
\affiliation[3]{THU}
\affiliation[4]{Samsung Research}
\affiliation[5]{CASIA}
\affiliation[6]{CUHK}

\contribution[*]{Equal contribution.}
\contribution[\dagger]{Project Leader.}
\contribution[\ddagger]{Corresponding authors.}

\correspondence{\email{hankyang428@gmail.com}}
\date{October 2026}
\checkdata[Code]{\url{https://github.com/Hhankyangg/envsched}}

\abstract{
As executable environments for language agents scale, generating and verifying their rollouts becomes increasingly expensive. Under finite rollout budgets, this raises an important question: can data management at the environment level make agentic reinforcement learning (RL) more efficient by organizing, evaluating, and scheduling task instances together with the environments that determine their tools, states, transitions, and verification? To address this question, we introduce \envsched (\textbf{Env}ironment-aware \textbf{Sched}uling), an online scheduler that learns from rollout history.
\envsched assigns scheduling value through two complementary modeling choices: exploitation prioritizes tasks likely to provide informative reward differences, while coverage spreads each batch across underobserved environments and interaction patterns. \envsched organizes environment--task pairs as nodes in a dual-relation graph, with within-environment edges capturing task semantics and shared tool, state, and verifier structure, and cross-environment edges capturing abstract execution and reward topology. By propagating historical rollout statistics along both edge types, it predicts within-task reward variance. These predictions guide a scheduler that allocates each batch to tasks with high predicted reward variance and underobserved environments and tasks. Environment quotas and a similarity penalty further limit concentration and redundancy, yielding informative and diverse batches without additional pilot rollouts. In RL on Qwen3-8B, \envsched achieves an average score of 48.81 across BFCL-v3, $\tau$-Bench, and ACEBench-Agent after 40 training steps, exceeding the highest score attained by random task selection after 120 steps. It also improves downstream performance across model sizes at the same training budget. Statistical analyses and ablation studies highlight the complementary roles of within-task reward differentiation and cross-environment coverage in these improvements.
}

\begin{document}
\maketitle
\pagestyle{plain}

\section{Introduction}
\label{sec:introduction}

\begin{figure*}[t]
\centering
\includegraphics[width=0.95\textwidth]{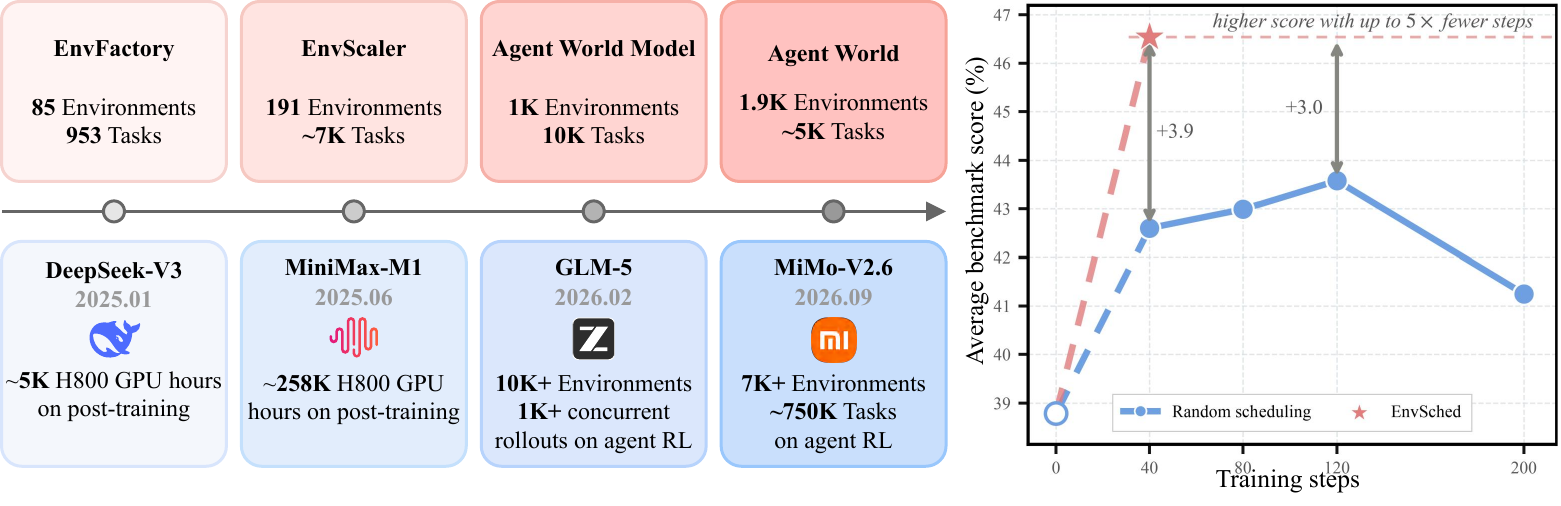}
\vspace{-2mm}
\caption{The left panel summarizes the growth of environments and RL resources in recent agent systems. The right panel shows \envsched's training-efficiency gain on Qwen3-8B under matched rollout budgets. Scores are averaged across BFCL-v3, $\tau$-Bench, and ACEBench-Agent.}
\vspace{-4mm}
\label{fig:intro}
\end{figure*}

Large language models (LLMs) are increasingly trained to act in stateful tool environments, where completing a user request requires multi-turn interaction, coordinated tool use, and consistent updates to the underlying state~\citep{bai2026clawgym,li2026agentic,ye2026claw}. 
For example, an agent rescheduling a reservation must retrieve the relevant records, check availability and domain rules, and update dependent entries in a valid order. Executable environments support this interaction through database-backed states, tool interfaces, transition logic, and task-specific verifiers. 
Recent systems have made executable environments a central scaling axis for Agentic RL, expanding training pools from tens to thousands of environments~\citep{xu2026envfactory,song2026envscaler,wang2026awm,dong2026agentworld}. This expansion comes with substantial training costs: industrial reports describe RL over thousands of environments and post-training runs consuming hundreds of thousands of GPU-hours~\citep{liu2024deepseek,chen2025minimax,zeng2026glm,mimo2026v26pro}. Figure~\ref{fig:intro} (left) summarizes the growing scale of environments and reported RL resources.

As the number of executable environments grows, the cost of generating and verifying RL rollouts grows with them. 
Researchers~\citep{zhang2025landscape,lyu2026mock} and industrial technical reports~\citep{zeng2026glm,mimo2026v26pro} increasingly emphasize that training efficiency must be improved through better organization of the training data itself, including environment design, trajectory generation, rollout scaling, and curriculum learning. Data-centric systems such as LLM-based subpopulation analysis and unified data workflow automation further illustrate the value of structuring and auditing data before model training~\citep{luo2024llm,liang2025dataflow}. 
Existing work has begun to study online task selection during RL, estimating training value from task difficulty, prompt representations, or historical rewards~\citep{sun2025dots,mao2026dynamics,liang2026leeps}. 
However, these methods model the prompt or task while leaving the executable environment outside the selection model. This misses a key source of variation in Agentic RL: the environment determines the tools, initial state, transition rules, and verifier, and thus shapes how a task unfolds and what learning signal it produces. This motivates our formulation of environment-level data management.

We formulate this challenge as \emph{environment-level data management}: organizing, evaluating, and scheduling task instances together with the executable environments in which they are carried out. Each environment specifies the available tools, initial and intermediate states, transition rules, and task verifier.
To address this challenge, we draw on the update mechanism of group-relative RL methods such as GRPO~\citep{shao2024deepseekmath}. In these methods, reward differences among rollouts of the same task determine their relative advantages; tasks with nearly identical outcomes therefore provide little learning signal. The scheduler should prioritize tasks whose current reward distributions are likely to yield informative differences. At the same time, repeatedly training on a small set of environments restricts the tools, states, transitions, and verification patterns seen by the policy, even when the selected tasks are individually informative. Effective scheduling must therefore preserve diversity across environments and interaction patterns.
Building on these two requirements, we introduce \envsched (\textbf{Env}ironment-aware \textbf{Sched}uling), an online scheduler learning from rollout history. It prioritizes task instances with high predicted reward informativeness, while it also spreads each batch across underobserved environments and structurally distinct interactions.

\envsched implements exploitation and coverage as two coordinated parts of the same scheduling procedure.
To support both decisions, \envsched represents environment--task pairs as nodes in a dual-relation graph. Within-environment edges connect tasks through shared semantics, tools, database fields, and verifier constraints. Cross-environment edges connect tasks through abstract execution and reward topology, enabling information transfer across similar environments. 
For example, reassigning a consultant and moving a patient between hospital beds can share a dependency structure of entity lookup, eligibility checking, relation replacement, and dependent-state validation. 

At each training step, \envsched aggregates historical rollout statistics separately along the two edge types and combines the resulting estimates to predict each task's reward variance under the evolving policy. It then allocates half of each batch to underobserved environments and tasks for coverage, and half to tasks with high predicted reward variance for exploitation. Environment quotas and a similarity penalty limit concentration and redundancy within the shared batch, while newly collected rollout outcomes update the node histories for subsequent rounds. This rollout--update--propagation--selection loop balances policy-dependent task prioritization with explicit environment coverage without additional pilot rollouts.

We evaluate \envsched by conducting reinforcement learning on the EnvScaler training pool with matched per-step rollout budgets. The right side of Figure~\ref{fig:intro} shows that \envsched reaches an average score of 48.81 on Qwen3-8B across BFCL-v3, $\tau$-Bench, and ACEBench-Agent~\citep{patil2025bfcl,yao2025taubench,chen2025acebench} after 40 steps, exceeding the highest score attained by random task selection during 120 steps. It also improves downstream performance at the same training budget, with gains across model sizes. Analysis of selected-task rollouts shows that mean within-task reward variance rises from 0.03208 to 0.04290, and environments covered per batch also rises from 36.53 to 45.96 on Qwen3-4B. 
Ablation studies confirm that reward differentiation and environment coverage are complementary: either component alone is insufficient for effective training. 
More broadly, \envsched demonstrates the distinctive value to efficient Agentic RL training, paving the way for scaling increasingly capable agent models with better use of training budgets.

Our contributions are threefold:
\begin{itemize}
    \item We are the first to formulate environment-level data management for Agentic RL, jointly modeling task specifications and executable environment structure.
    \item We propose \envsched, an Agentic RL training data scheduler which combines a dual-relation graph, historical reward-variance prediction, and coverage-aware batch construction to schedule training tasks without additional pilot rollouts.
    \item We demonstrate consistent gains in rollout efficiency across models of varying sizes and improved performance on a range of downstream tool-use benchmarks. Our analyses show that reward differentiation and environment coverage are complementary sources of training value, establishing environment-level data management as a promising data-centric direction for scaling efficient Agentic RL.
\end{itemize}

\section{Method}
\label{sec:method}

\subsection{Problem Setup and Training Loop}
\label{sec:method-setup}

\envsched treats an environment--task pair as the basic unit of training and scheduling. An environment is an executable setting that defines the tool interfaces, transition rules, state representation, and verifier; a task instance specifies the request and its initial state within that environment. Each pair therefore identifies one concrete rollout problem, and the model interacts with it through text responses and tool calls to produce a trajectory and its task-specific evaluation reward. At each training step, the scheduler chooses a batch of distinct pairs for rollout and model update.

\begin{figure*}[t]
\centering
{\includegraphics[width=0.95\textwidth]{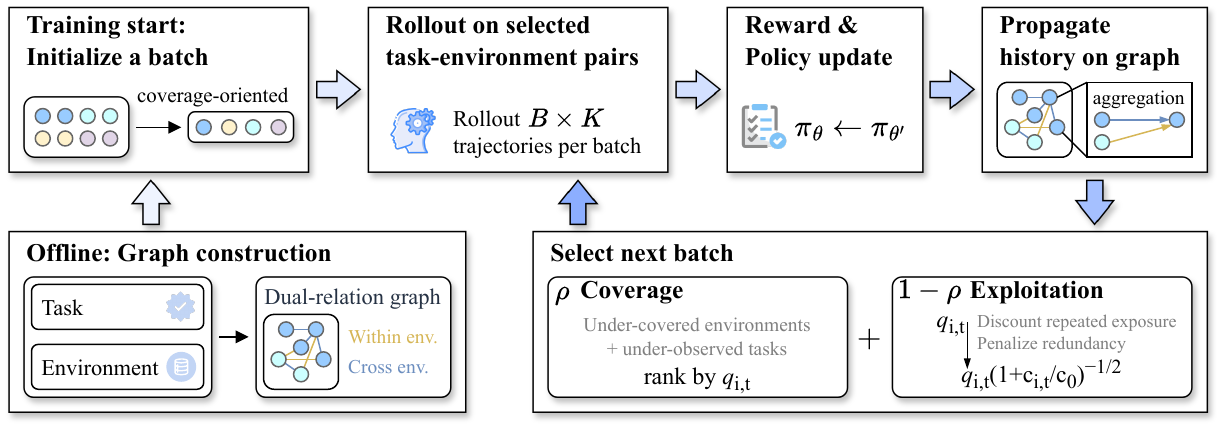}}
\vspace{-1mm}
\caption{\envsched training loop. Circles with the same color denote task--environment pairs from the same environment.}
\vspace{-5mm}
\label{fig:training-flow}
\end{figure*}

As illustrated in Figure~\ref{fig:training-flow}, \envsched builds a dual-relation graph from task specifications and executable environment structure before training, then initializes the schedule with a coverage-oriented batch. Rollouts on the selected pairs produce rewards for model updates and statistics for the corresponding node histories. Before the next training step, historical statistics propagate along within-environment and cross-environment edges to estimate task informativeness. The scheduler combines coverage of underobserved environments and tasks with exploitation of informative tasks, using environment quotas and a similarity penalty to limit concentration and redundancy. This rollout--update--propagation--selection loop repeats throughout training.

\subsection{dual-relation graph}
\label{sec:dual-relation-graph}

We define $\mathcal{G}=(\mathcal{V},\mathcal{E}^{\mathrm{in}},\mathcal{E}^{\mathrm{cross}})$ over the environment--task pairs. Each node stores the pair's historical reward statistics and selection count. The two edge types are constructed from similarity: within-environment similarities compare task semantics and concrete tool, state, and verifier features among pairs in the same environment; cross-environment similarities compare abstract execution and reward structure among pairs in different environments. Similarity determines which observed nodes contribute evidence and how strongly they contribute to the target node's estimate.

Before training, a language model reads the task instruction, initial-state summary, tool descriptions, and verifier to extract a structured operation description. It records required operations, prerequisites, execution dependencies, and relationships between operations and reward checks. For within-environment similarity, six TF--IDF views encode instruction words and character fragments, verifier-check text, verifier code structure, initial context, and the complete operation description. For cross-environment similarity, tool names, field names, entity identifiers, and target values are removed, and operations and conditions are mapped to categories such as lookup, eligibility checking, relation replacement, and dependent-state validation. The LLM prompt is provided in Appendix~\ref{app:topology-extraction}. Let $h_i^{(a)}$ encode concrete view $a$ and $g_i$ encode the abstract description. The similarities are

\begin{equation}
S_{ij}^{\mathrm{full}}=\sum_{a=1}^{6}\alpha_a\cos(h_i^{(a)},h_j^{(a)}),
\qquad S_{ij}^{\mathrm{abs}}=\cos(g_i,g_j).
\label{eq:task-similarities}
\end{equation}

The full similarity retains domain-specific information, while the abstract similarity supports transfer between tasks with different business terminology. For example, consultant reassignment and hospital-bed transfer can involve similar lookup, eligibility, and relation-update operations. At each training step, each target node selects its most similar observed neighbors separately within and across environments, using $S^{\mathrm{full}}$ and $S^{\mathrm{abs}}$, respectively. We denote these sets by $\mathcal{N}_t^{\mathrm{in}}(i)$ and $\mathcal{N}_t^{\mathrm{cross}}(i)$, with $|\mathcal{N}_t^{\mathrm{in}}(i)|\le K_{\mathrm{within}}$ and $|\mathcal{N}_t^{\mathrm{cross}}(i)|\le K_{\mathrm{cross}}$. Similarities remain fixed during training; node histories and the available observed neighbors evolve. Feature weights and neighbor configurations are specified in Appendix~\ref{app:implementation}. Figure~\ref{fig:graph-method}(a) summarizes this construction: colored regions group task--environment pairs by environment, while the two edge types transfer historical reward information through concrete and abstract structural similarities.

Figure~\ref{fig:graph-method}(b) summarizes how observed nodes inform estimates for the pool before the next rollout step: the two relation-specific estimates form a graph prior, which is fused with the target's own history and then calibrated into the informativeness score. Each observation records the mean reward $m$, the within-task reward variance $v$, and the mean-corrected dispersion $d$. Let $K$ denote the number of rollouts per task and $r_{i,s,1},\ldots,r_{i,s,K}$ the rewards of the complete rollout group for task $i$ at step $s$; we compute the population variance over all $K$ rewards:

\begin{equation}
m_{i,s}=\frac{1}{K}\sum_{k=1}^{K}r_{i,s,k},
\qquad
v_{i,s}=\operatorname{Var}_{k=1,\ldots,K}(r_{i,s,k})
=\frac{1}{K}\sum_{k=1}^{K}\left(r_{i,s,k}-m_{i,s}\right)^2.
\label{eq:group-variance}
\end{equation}

This statistic directly measures the reward differentiation available within the group used for the policy update. The mean reward and mean-corrected dispersion provide complementary information about the same group-level reward distribution; their definitions appear in Appendix~\ref{app:history-estimation}.

\begin{figure*}[t]
\centering
{\includegraphics[width=0.95\textwidth]{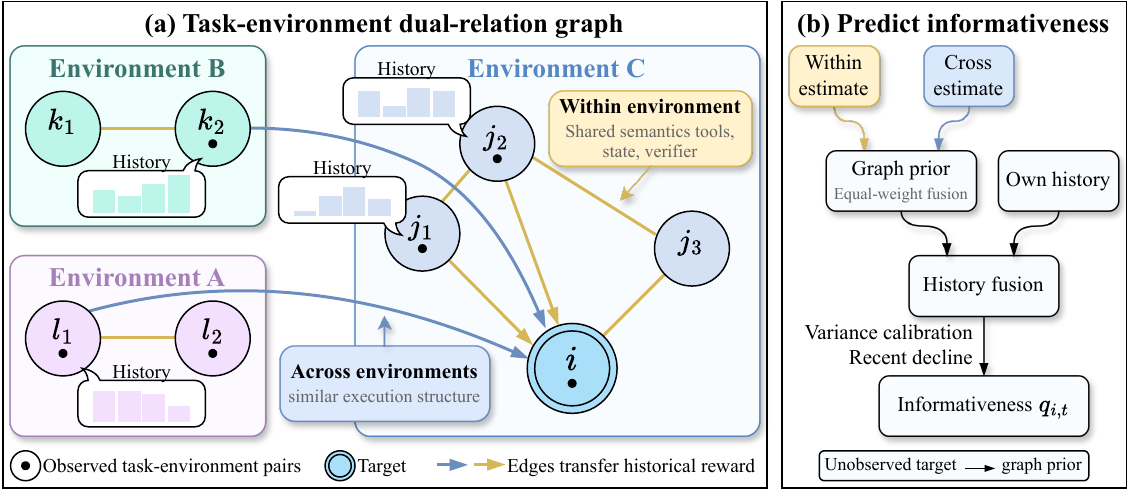}}
\caption{Dual-relation graph and informativeness prediction. (a) Within-environment and cross-environment edges propagate historical statistics from observed task--environment pairs to the target. (b) Relation-specific estimates are fused with the target's history to produce the final score.}
\vspace{-5mm}

\label{fig:graph-method}
\end{figure*}

\subsection{Predicting Task Informativeness}
\label{sec:task-informativeness}

For each statistic $y\in\{v,m,d\}$, we discount older observations and center them against their step-level mean before shifting to the current global reference $b_t^y$. Let $\bar y_{i,t}$ denote the historical estimate and $W_{i,t}$ its effective observation weight. Along each relation $r\in\{\mathrm{in},\mathrm{cross}\}$, we aggregate observed neighbors with weights $a_{ij,t}^r$ determined by similarity and the amount of recent history:

\begin{equation}
G_{i,t}^{r,y}=\frac{\sum_{j\in\mathcal{N}_t^r(i)}a_{ij,t}^r\bar y_{j,t}+\lambda_g b_t^y}
{\sum_{j\in\mathcal{N}_t^r(i)}a_{ij,t}^r+\lambda_g}.
\label{eq:typed-propagation}
\end{equation}

Shrinkage toward $b_t^y$ stabilizes estimates with little neighbor evidence. We weight the two relation types equally and average their estimates into a graph prior, then combine it with the target's history:

\begin{equation}
P_{i,t}^{y}=\tfrac12 G_{i,t}^{\mathrm{in},y}+\tfrac12 G_{i,t}^{\mathrm{cross},y},
\qquad \widehat y_{i,t}=\frac{W_{i,t}\bar y_{i,t}+\lambda_s P_{i,t}^{y}}{W_{i,t}+\lambda_s}.
\label{eq:history-fusion}
\end{equation}

An unobserved node relies on the graph prior; its own observations gain weight as its history grows. We combine the predicted variance with a variance estimate reconstructed from the predicted mean and dispersion, then compare estimates from long and short history windows. A recent decline lowers the priority. The resulting percentile score $q_{i,t}$ ranks tasks for scheduling. Appendix~\ref{app:history-estimation} gives the complete calibration, fusion, and scoring definitions.

\subsection{Coverage-Aware Task Scheduling}
\label{sec:coverage-scheduling}

The scheduler assigns a fraction $\rho$ of each batch to coverage and the remaining fraction $1-\rho$ to exploitation. Coverage first restricts the candidate set using a lexicographic priority over current-batch environment coverage, cumulative environment exposure, and task observation count. Within the resulting candidate set, it prioritizes predicted informativeness. Exploitation prioritizes predicted informativeness while gradually discounting repeated exposure. For cumulative selection count $c_{i,t}$, the exposure factor is $\omega_{i,t}=(1+c_{i,t}/c_0)^{-1/2}$, and the exploitation score is $q_{i,t}\omega_{i,t}$.

Both phases build a shared batch greedily. Let $B_{\mathrm{cov}}=\lfloor\rho B\rfloor$ and $B_{\mathrm{exp}}=B-B_{\mathrm{cov}}$ be the phase budgets. Given the eligible candidates $\mathcal{A}$ and the current selected set $\mathcal{B}$, the next node is

\begin{equation}
i^*=\arg\max_{i\in\mathcal{A}}\left[a_{i,t}^{\mathrm{phase}}-\lambda_{\mathrm{sim}}\max_{j\in\mathcal{B}}S_{ij}^{\mathrm{full}}\right],
\label{eq:batch-selection}
\end{equation}

where the phase score is $q_{i,t}$ for coverage and $q_{i,t}\omega_{i,t}$ for exploitation. For coverage, $\mathcal{A}$ is first restricted by the lexicographic coverage priority described above; Eq.~\ref{eq:batch-selection} then resolves the remaining choices using informativeness and similarity. The similarity penalty is zero for an empty batch. It uses the full task representation across selected pairs to reduce redundant batch content. Environment quotas and consecutive-selection limits further control concentration.

Candidate eligibility is part of the scheduler and includes two additional filters. A progress filter temporarily excludes tasks with non-positive estimated progress from exploitation while retaining coverage opportunities for reevaluation. An identifier filter uses fixed-action replay to confirm suspicious reward-producing actions and removes the affected tasks from both phases. These filters change candidate eligibility rather than the ranking score. At initialization, the scheduler selects environment representatives and fills the remaining slots using task novelty. The exact eligibility rules and initialization procedure appear in Appendix~\ref{app:scheduling-rules}.

\section{Experiments}

\begin{table*}[t]
\caption{Main results. ``Overall'' refers to the weighted score. \textcolor{deepgreen}{Green} and \textcolor{deepred}{red} subscripts denote variations relative to random scheduling.}
\label{tab:main_results}
\centering

\resizebox{\textwidth}{!}{
\renewcommand{\arraystretch}{0.8}

\setlength{\tabcolsep}{0.55mm}
\begin{tabular}{lcccccccccccc}
\toprule
\multirow{2}{*}{\textbf{Model}} & \multirow{2}{*}{\textbf{Steps}} & \multicolumn{5}{c}{\textbf{BFCL-v3 Multi-Turn}} & \multicolumn{3}{c}{\textbf{$\tau$-Bench}} & \multicolumn{3}{c}{\textbf{ACEBench-Agent}} \\
\cmidrule(r){3-7}\cmidrule(lr){8-10}\cmidrule(lr){11-13}
 && \textbf{Base}& \textbf{\makecell{Miss-\\Func}} & \textbf{\makecell{Miss-\\Param}} & \textbf{\makecell{Long-\\Context}} & \textbf{Overall} & \textbf{Retail} & \textbf{Airline} & \textbf{Overall} & \textbf{\makecell{Multi-\\Step}} & \textbf{\makecell{Multi-\\Turn}} & \textbf{Overall} \\
\midrule
\rowcolor[gray]{0.95} Qwen3-1.7B &-& 13.50 & 6.00 & 12.50 & 7.00 & \textbf{9.75} & 8.99 & 16.00 & \textbf{11.11} & 35.00 & 27.78 & \textbf{30.67} \\
 \quad + Random & 200
& 13.17 & 10.50 & 10.50 & 7.67 & \textbf{10.46} & 15.36 & 16.00 & \textbf{15.56} & 35.00 & 24.44 & \textbf{28.67} \\
\rowcolor[gray]{0.95} \quad + DPS & 200
& 12.50 & 10.83 & 9.33 & 7.00 & \textbf{$\text{9.92}_{\textcolor{deepred}{\text{-0.54}}}$} & 16.81 & 17.33 & \textbf{$\text{16.97}_{\textcolor{deepgreen}{\text{+1.41}}}$} & 33.33 & 26.67 & \textbf{$\text{29.33}_{\textcolor{deepgreen}{\text{+0.66}}}$} \\
 \quad + LEEPS & 200
& 13.83 & 11.67 & 11.00 & 8.50 & \textbf{$\text{11.25}_{\textcolor{deepgreen}{\text{+0.79}}}$} & 16.52 & 17.33 & \textbf{$\text{16.77}_{\textcolor{deepgreen}{\text{+1.21}}}$} & 35.00 & 27.78 & \textbf{$\text{30.67}_{\textcolor{deepgreen}{\text{+2.00}}}$} \\

\rowcolor{envschedblue}\quad + \envsched  & 40
& 13.50 & 13.00 & 11.50 & 9.00 & \textbf{$\text{11.75}_{\textcolor{deepgreen}{\text{+1.29}}}$} & 17.10 & 16.67 & \textbf{$\text{16.97}_{\textcolor{deepgreen}{\text{+1.41}}}$} & 40.00 & 24.44 & \textbf{$\text{30.67}_{\textcolor{deepgreen}{\text{+2.00}}}$} \\

\rowcolor{envschedblue}\quad + \envsched  & 200
& 14.00 & 14.50 & 12.17 & 9.00 & \textbf{$\text{12.42}_{\textcolor{deepgreen}{\text{+1.96}}}$} & 18.26 & 16.00 & \textbf{$\text{17.58}_{\textcolor{deepgreen}{\text{+2.02}}}$} & 41.67 & 24.44 & \textbf{$\text{31.33}_{\textcolor{deepgreen}{\text{+2.66}}}$} \\
\midrule
\rowcolor[gray]{0.95} Qwen3-4B&-& 32.00 & 20.00 & 24.00 & 25.50 & \textbf{25.38} & 40.87 & 26.00 & \textbf{36.36} & 56.67 & 51.11 & \textbf{53.33} \\
\quad + Random & 200
& 36.67 & 26.17 & 25.50 & 27.00 & \textbf{28.83} & 40.87 & 30.67 & \textbf{37.78} & 70.00 & 46.67 & \textbf{56.00} \\
\rowcolor[gray]{0.95} \quad + DPS & 200
& 38.33 & 27.00 & 26.67 & 29.50 & \textbf{$\text{30.38}_{\textcolor{deepgreen}{\text{+1.55}}}$} & 39.42 & 30.67 & \textbf{$\text{36.77}_{\textcolor{deepred}{\text{-1.01}}}$} & 70.00 & 47.78 & \textbf{$\text{56.67}_{\textcolor{deepgreen}{\text{+0.67}}}$} \\
 \quad + LEEPS & 200
& 38.00 & 27.00 & 26.33 & 28.67 & \textbf{$\text{30.00}_{\textcolor{deepgreen}{\text{+1.17}}}$} & 42.32 & 30.67 & \textbf{$\text{38.79}_{\textcolor{deepgreen}{\text{+1.01}}}$} & 70.00 & 47.78 & \textbf{$\text{56.67}_{\textcolor{deepgreen}{\text{+0.67}}}$} \\
\rowcolor{envschedblue}\quad + \envsched  & 40
& 36.00 & 27.67 & 26.50 & 30.50 & \textbf{$\text{30.17}_{\textcolor{deepgreen}{\text{+1.34}}}$} & 45.80 & 30.00 & \textbf{$\text{41.01}_{\textcolor{deepgreen}{\text{+3.23}}}$} & 70.00 & 50.00 & \textbf{$\text{58.00}_{\textcolor{deepgreen}{\text{+2.00}}}$} \\
\rowcolor{envschedblue}\quad + \envsched & 200
& 40.67 & 28.17 & 27.50 & 31.17 & \textbf{$\text{31.88}_{\textcolor{deepgreen}{\text{+3.05}}}$} & 45.51 & 30.67 & \textbf{$\text{41.01}_{\textcolor{deepgreen}{\text{+3.23}}}$} & 70.00 & 50.00 & \textbf{$\text{58.00}_{\textcolor{deepgreen}{\text{+2.00}}}$} \\
\midrule
\rowcolor[gray]{0.95} Qwen3-8B &-& 32.00 & 33.50 & 22.00 & 28.00 & \textbf{28.88} & 46.38 & 30.00 & \textbf{41.41} & 61.67 & 56.67 & \textbf{58.67} \\
\quad + Random & 200
& 37.50 & 34.00 & 27.50 & 36.50 & \textbf{33.88} & 56.23 & 26.67 & \textbf{47.27} & 61.67 & 51.11 & \textbf{55.33} \\
\rowcolor[gray]{0.95} \quad + DPS & 200
& 40.67 & 34.17 & 29.83 & 37.33 & \textbf{$\text{35.50}_{\textcolor{deepgreen}{\text{+1.62}}}$} & 57.68 & 26.00 & \textbf{$\text{48.08}_{\textcolor{deepgreen}{\text{+0.81}}}$} & 65.00 & 53.33 & \textbf{$\text{58.00}_{\textcolor{deepgreen}{\text{+2.67}}}$} \\
 \quad + LEEPS & 200
& 41.17 & 35.33 & 29.33 & 37.00 & \textbf{$\text{35.71}_{\textcolor{deepgreen}{\text{+1.83}}}$} & 56.23 & 28.00 & \textbf{$\text{47.68}_{\textcolor{deepgreen}{\text{+0.41}}}$} & 60.00 & 51.11 & \textbf{$\text{54.67}_{\textcolor{deepred}{\text{-0.66}}}$} \\
\rowcolor{envschedblue}\quad + \envsched  & 40
& 45.17 & 35.83 & 31.50 & 38.33 & \textbf{$\text{37.71}_{\textcolor{deepgreen}{\text{+3.83}}}$} & 58.26 & 33.33 & \textbf{$\text{50.71}_{\textcolor{deepgreen}{\text{+3.44}}}$} & 70.00 & 50.00 & \textbf{$\text{58.00}_{\textcolor{deepgreen}{\text{+2.67}}}$} \\
\rowcolor{envschedblue}\quad + \envsched & 200
& 46.50 & 35.00 & 31.00 & 39.00 & \textbf{$\text{37.88}_{\textcolor{deepgreen}{\text{+4.00}}}$} & 58.26 & 32.00 & \textbf{$\text{50.30}_{\textcolor{deepgreen}{\text{+3.03}}}$} & 70.00 & 51.11 & \textbf{$\text{58.67}_{\textcolor{deepgreen}{\text{+3.34}}}$} \\
\midrule
\rowcolor[gray]{0.95} Qwen3-30B-A3B & -
& 39.50 & 27.00 & 33.00 & 32.00 & \textbf{32.88}
& 61.45 & 32.00 & \textbf{52.53}
& 80.00 & 56.67 & \textbf{66.00} \\

\quad + Random & 200
& 40.67 & 29.17 & 30.00 & 35.67 & \textbf{33.88}
& 57.39 & 38.67 & \textbf{51.72}
& 81.67 & 58.89 & \textbf{68.00} \\

\rowcolor{envschedblue}\quad + \envsched & 40
& 43.00 & 32.17 & 28.17 & 36.67 & \textbf{$\text{35.00}_{\textcolor{deepgreen}{\text{+1.13}}}$}
& 60.00 & 38.67 & \textbf{$\text{53.54}_{\textcolor{deepgreen}{\text{+1.82}}}$}
& 86.67 & 58.89 & \textbf{$\text{70.00}_{\textcolor{deepgreen}{\text{+2.00}}}$} \\

\rowcolor{envschedblue}\quad + \envsched & 200
& 47.00 & 35.67 & 30.00 & 38.50 & \textbf{$\text{37.79}_{\textcolor{deepgreen}{\text{+3.92}}}$}
& 60.00 & 42.67 & \textbf{$\text{54.75}_{\textcolor{deepgreen}{\text{+3.03}}}$}
& 86.67 & 60.00 & \textbf{$\text{70.67}_{\textcolor{deepgreen}{\text{+2.67}}}$} \\

\bottomrule
\end{tabular}
}
\vspace{-4mm}
\end{table*}

\subsection{Experimental Setup}
\label{sec:experimental-setup}
\textbf{Training Environments and Evaluation.}
We train Qwen3-Thinking~\citep{yang2025qwen3} models with GRPO~\citep{shao2024deepseekmath} on the EnvScaler~\citep{song2026envscaler} executable-environment pool, which contains 2,550 task instances across 51 executable environments. Each task instance is evaluated through its environment-specific tools, initial state, and terminal verifier. We report performance on BFCL-v3 Multi-Turn, $\tau$-Bench, and ACEBench-Agent
~\citep{patil2025bfcl,yao2025taubench,chen2025acebench}, covering function calling, multi-turn interaction, and long-horizon tool use.

\textbf{Implementation Details.}
For training with \envsched, we apply a unified GRPO configuration across Qwen3-1.7B, Qwen3-4B, Qwen3-8B, and Qwen3-30B-A3B. Each training batch contains $B=64$ environment--task pairs with rollout $K=8$ per pair (512 trajectories). The scheduler uses a fixed coverage fraction $\rho=1/2$, yielding 32 coverage pairs and 32 exploitation pairs. We train for 200 single-epoch steps using Adam ($\text{lr}=10^{-6}$, cosine schedule, 10 warmup steps) with a policy-ratio clip of 0.2 and batch-whitened, $[-0.2,0.2]$-clipped advantages. Rollouts sample at temperature 0.99 (top-$p=0.99$, top-$k=100$) with limits of 40 turns, 4,096 tokens per turn, and 32,000 tokens per sequence. The policy loss covers only model-generated tokens, including reasoning; all other tokens are masked. All scheduler and optimization hyperparameters are fixed across model sizes. The default graph neighborhood uses $K_{\mathrm{within}}=6$ within-environment neighbors and $K_{\mathrm{cross}}=12$ cross-environment neighbors. Graph representations, scheduling constants, and remaining hyperparameters are listed in Appendix~\ref{app:implementation}.
For testing on the benchmarks, we use GPT-5.6-terra~\citep{singh2025openai} as the user model for $\tau$-Bench, and GPT-4o-2024-08-06~\citep{hurst2024gpt} for ACEBench-Agent. We repeat all experiments with three random seeds and report seed-aggregated results. The Qwen3-30B-A3B experiment uses the same task pool, rollout budget, scheduler and optimization hyperparameters, benchmark protocol, and three-seed evaluation. We compare the untrained model, random scheduling at 200 steps, and \envsched at 40 and 200 steps.

\textbf{Baselines.}
The main comparison includes base model without training, random scheduling, and \envsched, along with recent related training data scheduling methods DPS~\citep{mao2026dynamics} and LEEPS~\citep{liang2026leeps}. 
Random samples task--environment pairs uniformly from the training pool. 
DPS (Dynamics-Predictive Sampling) models prompt-solving dynamics from historical rollout outcomes and selects prompts predicted to provide informative learning signals before rollout generation.
LEEPS combines representation-space neighbors, historical rollout outcomes, and explore--exploit portfolios to select prompts likely to yield nonzero reward variance.

\subsection{Main Results}

\begin{figure*}[t]
\centering
\includegraphics[width=\textwidth]{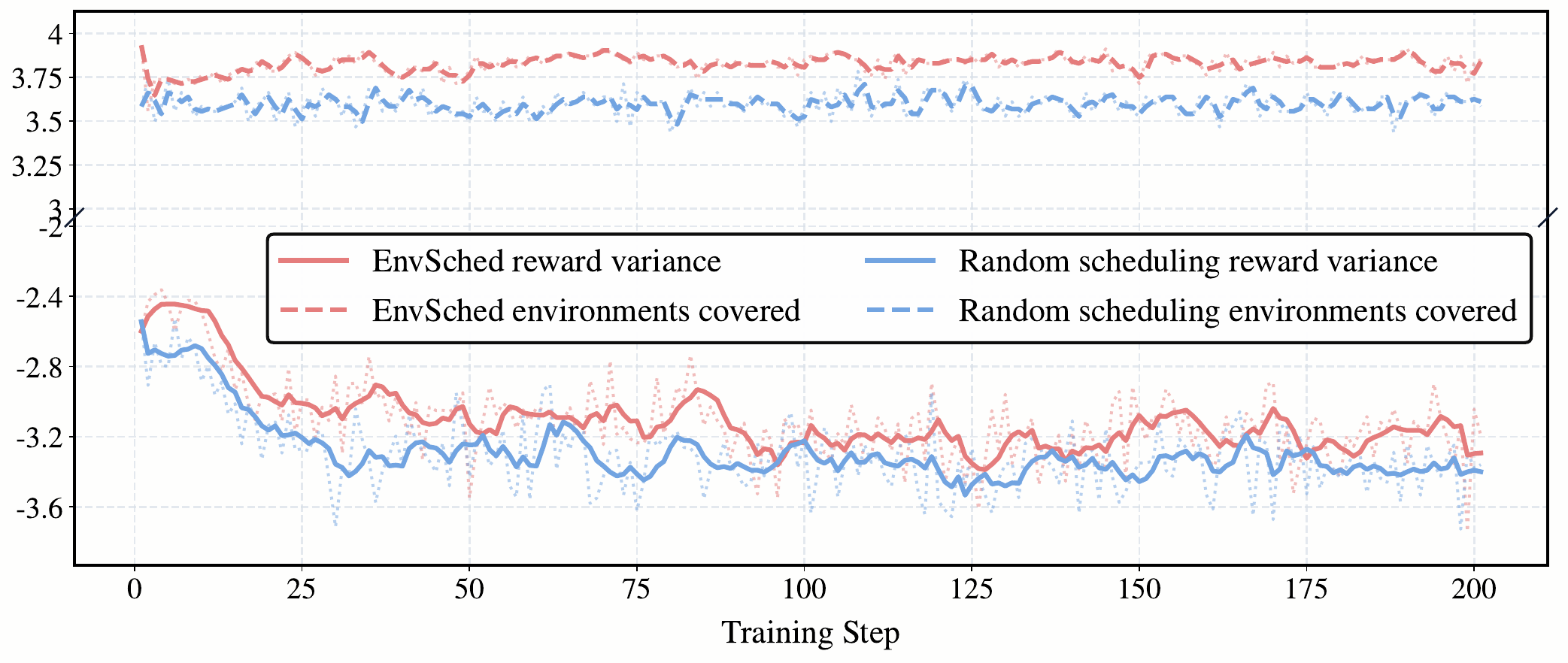}
\caption{Selected-task reward variance and environment coverage over training steps, with log-transformed vertical-axis values.}
\label{fig:reward-env}
\vspace{-5mm}
\end{figure*}

\begin{table*}[t]
\caption{Training statistics for random scheduling and \envsched over 200 GRPO training steps. BFCL-v3 Multi-Turn reports the overall score.}
\label{tab:analysis}
\centering
\resizebox{\textwidth}{!}{
\renewcommand{\arraystretch}{0.8}

\begin{tabular}{lcccccc}
\toprule
& \makecell{\textbf{Mean reward}\\\textbf{variance}}
& \makecell{\textbf{Median reward}\\\textbf{variance}}
& \makecell{\textbf{Median}\\\textbf{reward}}
& \makecell{\textbf{Environments covered}\\\textbf{per batch}}
& \makecell{\textbf{Tools used}\\\textbf{per batch}}
& \makecell{\textbf{BFCL-v3}\\\textbf{Multi-Turn}} \\
\midrule
Random & 0.03208 & 0.01468 & 0.595 & 36.53  & 413.41 & 28.83 \\
\envsched & 0.04290 & 0.02108 & 0.592 & 45.96 & 478.39 & 31.88 \\
\midrule
\rowcolor{envschedblue} \textbf{Difference} & \textcolor{deepgreen}{\text{+34\%}} & \textcolor{deepgreen}{\text{+44\%}} & \textcolor{deepred}{\text{-0.5\%}} & \textcolor{deepgreen}{\text{+26\%}} & \textcolor{deepgreen}{\text{+16\%}} & \textcolor{deepgreen}{\text{+3.05}} \\
\bottomrule
\end{tabular}
}
\vspace{-2mm}
\end{table*}

We implement all baselines against the same EnvScaler pool and enforce the same batch size, rollout count, training horizon, and model-update procedure as \envsched. The results on the benchmarks are presented in Table~\ref{tab:main_results}. In addition to the results after 200 training steps, we also include the performance of \envsched after just 40 steps. Next, we analyze the results from three aspects:

\textbf{Consistent gains across model sizes at 200 steps.}
At the 200-step budget, \envsched improves over random scheduling for all three benchmark overall scores at every model size. Qwen3-4B illustrates the pattern clearly: BFCL-v3 rises from 28.83 to 31.88 (+3.05), $\tau$-Bench from 37.78 to 41.01 (+3.23), and ACEBench-Agent from 56.00 to 58.00 (+2.00). The gains also hold for the other dense models: at 1.7B, the improvements are +1.96, +2.02, and +2.66, while at 8B they are +4.00, +3.03, and +3.34 on BFCL-v3, $\tau$-Bench, and ACEBench-Agent, respectively. The BFCL gain at 4B is distributed across all four categories, including Base (36.67 to 40.67), Miss-Function (26.17 to 28.17), Miss-Parameter (25.50 to 27.50), and Long-Context (27.00 to 31.17). The gains extend to Qwen3-30B-A3B under the same configuration: BFCL-v3, $\tau$-Bench, and ACEBench-Agent reach 37.79, 54.75, and 70.67, improving over random scheduling by 3.92, 3.03, and 2.67 points, respectively, without model-specific scheduler tuning.

\textbf{High efficiency.}
\envsched is already effective after 40 steps within the 200-step training budget. At 4B, it reaches 30.17 on BFCL-v3, 41.01 on $\tau$-Bench, and 58.00 on ACEBench-Agent, matching its 200-step ACEBench-Agent score. At 8B, the 40-step $\tau$-Bench score reaches 50.71, above the 200-step score of 50.30; at 1.7B, ACEBench-Agent reaches 30.67 versus 31.33 after 200 steps. On Qwen3-30B-A3B, the 40-step scores of 35.00, 53.54, and 70.00 exceed the 200-step random baseline on all three benchmarks. These results show that \envsched can extract useful training signal quickly and, on some benchmarks, match or exceed its longer-run performance.

\textbf{Prompt-only schedulers do not transfer reliably.}
DPS and LEEPS are designed around the informativeness of individual prompts, without modeling the environment coupled to each task. Here, choosing a prompt also chooses the tools, state, and constraints exposed to the policy, so prompt-level selection alone cannot reliably identify useful training data. Both methods underperform \envsched overall, and some of their dimensions are no better than random: DPS is below random on 4B $\tau$-Bench (36.77 vs.~37.78), and LEEPS on 8B ACEBench-Agent (54.67 vs.~55.33). The comparison supports combining reward differentiation with environment-aware coverage.

\subsection{Analysis}
\subsubsection{Reward Variance and Environment Coverage}



On Qwen3-4B, \envsched improves both reward differentiation and coverage under the same 64-task batch budget (Table~\ref{tab:analysis}). Compared with random scheduling, it increases mean within-task reward variance from 0.03208 to 0.04290 (+34\%) and median variance by 44\%, while median reward remains nearly unchanged. Environment coverage rises from 36.53 to 45.96 per batch (+26\%) and tool coverage from 413.41 to 478.39 (+16\%), alongside a BFCL-v3 improvement from 28.83 to 31.88. The log-transformed curves in Figure~\ref{fig:reward-env} show higher variance and coverage for most of the 200 training steps. 
Over training, variance declines, but the relative magnitude relationships are preserved, while coverage stays stable, preserving reward differentiation and environment diversity.

\begin{table*}[t]
\caption{Component ablations and neighborhood-size sensitivity. Panel (a) varies one component of the full scheduler; panel (b) varies one neighbor count at a time in complete 200-step training runs. Benchmark columns report weighted scores. Green and red subscripts denote changes relative to the full \envsched or the default neighborhood size.}
\label{tab:ablation-components-topk}
\centering
\begingroup
\small
\setlength{\tabcolsep}{4pt}
\renewcommand{\arraystretch}{0.8}
\resizebox{\textwidth}{!}{%
\begin{tabular}{lcccccc}
\toprule
\textbf{Configuration} & \makecell{\textbf{BFCL-v3}\\\textbf{Multi-Turn}} & \textbf{$\tau$-Bench} & \makecell{\textbf{ACEBench}\\\textbf{Agent}} & \makecell{\textbf{Mean reward}\\\textbf{variance}} & \makecell{\textbf{Environments}\\\textbf{per batch}} & \makecell{\textbf{Tools used}\\\textbf{per batch}} \\
\midrule
\rowcolor[gray]{0.95} Random  & 28.83 & 37.78 & 56.00 & 0.0321 & 36.53 & 413.41 \\
\midrule
\multicolumn{7}{l}{\textbf{(a) Component contributions}} \\
\rowcolor{envschedblue} Full \envsched & 31.88 & 41.01 & 58.00 & 0.0429 & 45.96 & 478.39 \\
Within-environment edges only & $\text{30.42}_{\textcolor{deepred}{\text{-1.46}}}$ & $\text{40.40}_{\textcolor{deepred}{\text{-0.61}}}$ & 58.00 & $0.0391_{\textcolor{deepred}{\text{-0.0038}}}$ & $46.70_{\textcolor{deepgreen}{\text{+0.74}}}$ & $498.55_{\textcolor{deepgreen}{\text{+20.16}}}$ \\
\rowcolor[gray]{0.95} Cross-environment edges only & $\text{30.63}_{\textcolor{deepred}{\text{-1.25}}}$ & 41.01 & $\text{57.33}_{\textcolor{deepred}{\text{-0.67}}}$ & $0.0383_{\textcolor{deepred}{\text{-0.0046}}}$ & $46.18_{\textcolor{deepgreen}{\text{+0.22}}}$ & $466.50_{\textcolor{deepred}{\text{-11.89}}}$ \\
Coverage only ($\rho=1$) & $\text{29.71}_{\textcolor{deepred}{\text{-2.17}}}$ & $\text{39.39}_{\textcolor{deepred}{\text{-1.62}}}$ & $\text{56.67}_{\textcolor{deepred}{\text{-1.33}}}$ & $0.0357_{\textcolor{deepred}{\text{-0.0072}}}$ & $51.00_{\textcolor{deepgreen}{\text{+5.04}}}$ & $502.10_{\textcolor{deepgreen}{\text{+23.71}}}$ \\
\rowcolor[gray]{0.95} Exploitation only ($\rho=0$) & $\text{29.17}_{\textcolor{deepred}{\text{-2.71}}}$ & $\text{38.59}_{\textcolor{deepred}{\text{-2.42}}}$ & $\text{56.00}_{\textcolor{deepred}{\text{-2.00}}}$ & $0.0442_{\textcolor{deepgreen}{\text{+0.0013}}}$ & $33.95_{\textcolor{deepred}{\text{-12.01}}}$ & $420.63_{\textcolor{deepred}{\text{-57.76}}}$ \\
\midrule
\multicolumn{7}{l}{\textbf{(b) Neighborhood size: $(K_{\mathrm{within}},K_{\mathrm{cross}})$}} \\
\rowcolor{envschedblue} Default $(6,12)$ & 31.88 & 41.01 & 58.00 & 0.0429 & 45.96 & 478.39 \\
Half within $(3,12)$ & $\text{31.25}_{\textcolor{deepred}{\text{-0.63}}}$ & $\text{41.62}_{\textcolor{deepgreen}{\text{+0.61}}}$ & $\text{56.67}_{\textcolor{deepred}{\text{-1.33}}}$ & $0.0409_{\textcolor{deepred}{\text{-0.0020}}}$ & $47.18_{\textcolor{deepgreen}{\text{+1.22}}}$ & $488.65_{\textcolor{deepgreen}{\text{+10.26}}}$ \\
\rowcolor[gray]{0.95} Double within $(12,12)$ & $\text{31.71}_{\textcolor{deepred}{\text{-0.17}}}$ & $\text{40.81}_{\textcolor{deepred}{\text{-0.20}}}$ & $\text{59.33}_{\textcolor{deepgreen}{\text{+1.33}}}$ & $0.0434_{\textcolor{deepgreen}{\text{+0.0005}}}$ & $45.66_{\textcolor{deepred}{\text{-0.30}}}$ & $475.71_{\textcolor{deepred}{\text{-2.68}}}$ \\
Half cross $(6,6)$ & $\text{30.92}_{\textcolor{deepred}{\text{-0.96}}}$ & $\text{40.20}_{\textcolor{deepred}{\text{-0.81}}}$ & $\text{57.33}_{\textcolor{deepred}{\text{-0.67}}}$ & $0.0425_{\textcolor{deepred}{\text{-0.0004}}}$ & $45.70_{\textcolor{deepred}{\text{-0.26}}}$ & $480.01_{\textcolor{deepgreen}{\text{+1.62}}}$ \\
\rowcolor[gray]{0.95} Double cross $(6,24)$ & $\text{32.83}_{\textcolor{deepgreen}{\text{+0.95}}}$ & $\text{41.01}$ & $\text{58.67}_{\textcolor{deepgreen}{\text{+0.67}}}$ & $0.0441_{\textcolor{deepgreen}{\text{+0.0012}}}$ & $46.90_{\textcolor{deepgreen}{\text{+0.94}}}$ & $493.83_{\textcolor{deepgreen}{\text{+15.44}}}$ \\
\bottomrule
\end{tabular}}
\endgroup
\end{table*}

\subsubsection{Computational Cost}
\envsched reuses historical training outcomes and requires zero additional probe trajectories. With cached static similarities and fixed-capacity neighbor heaps, an incremental scheduling step costs $O(N\Delta_t\log \hat{K}+N\hat{K}+NB)$, where $N$ is the task-pool size, $\Delta_t$ counts newly eligible neighbor candidates, $\hat{K}=K_{\mathrm{within}}+K_{\mathrm{cross}}$ is the neighbor budget, and $B$ is the batch size. For fixed $\Delta_t$, $\hat{K}$, and $B$, and without boundary-tie fallback scans, this cost is linear in $N$. In the measured 200-step run, online selection takes 6,985.94 seconds in total, or 34.93 seconds per step. This is a small fraction of the training step: a Qwen3-4B step takes approximately 25 minutes on H100 hardware, so scheduling accounts for about 2.3\% of the step time. The incremental backend uses $O(F+N\hat{K}+N)$ persistent space, where $F$ is the number of sparse feature entries. At the 2,550-task scale, sparse features occupy 85.87 MiB, neighbor and progress state occupies approximately 1.24 MiB, and the separately profiled two-matrix representation occupies 42.59 MiB on CPU. These bounded time and memory costs make online environment-aware scheduling practical within the training loop.

\subsection{Ablation Studies}
\subsubsection{Component Contributions}
We isolate the contributions of the two graph relations and the coverage--exploitation allocation. The relation ablations retain only within-environment edges or only cross-environment edges. The allocation ablations assign all 64 slots to coverage or all 64 slots to exploitation. Table~\ref{tab:ablation-components-topk} reports performance and statistics. 
The results show that reward differentiation and environment coverage must be optimized together. 
Coverage-only scheduling reaches the broadest batches (51.00 environments) but lowers mean reward variance to 0.0357 and degrades all three downstream scores, while exploitation-only scheduling produces the highest variance (0.0442) yet covers only 33.95 environments and performs substantially worse. 
Combining the two objectives recovers both informative reward differences and broad exposure. Full \envsched has the highest BFCL-v3 score among the component configurations and ties the within-environment-only variant on ACEBench-Agent. The cross-environment-only variant matches the full model on $\tau$-Bench but performs worse on BFCL-v3 and ACEBench-Agent. Both edge ablations also reduce mean reward variance relative to the full graph.
Modeling both relations therefore provides effective transfer of rollout evidence across tasks and environments; qualitative examples appear in Appendix~\ref{app:case-studies}.

\subsubsection{Initialization Robustness}
We examine sensitivity to the first batch with 100 independent seven-step training runs. Each run starts from a randomly sampled cold-start batch of 64 tasks and then follows the same 32/32 coverage--exploitation scheduler and training procedure. Table~\ref{tab:ablation-initialization} reports the mean and confidence interval across runs and compares the bottom and top quartiles ranked by the variance of the initial batch. The experiment measures how much random initialization affects the \envsched training.

\begin{table*}[t]
\caption{Initialization sensitivity over 100 seven-step runs. Panel (a) reports means with 95\% confidence intervals and 5th--95th percentile ranges; panel (b) compares the lowest and highest initial-variance quartiles. Hit rate measures exploitation tasks in the highest-variance quartile.}
\label{tab:ablation-initialization}
\centering
\begingroup
\small
\renewcommand{\arraystretch}{0.8}
\begin{tabular*}{\textwidth}{@{\extracolsep{\fill}}lccc@{}}
\toprule
\multicolumn{4}{l}{\textbf{(a) Selection quality across random initializations}} \\
\midrule
\textbf{Metric} & \textbf{Step 1} & \textbf{Step 7} & \textbf{Step 7 range} \\
\midrule
\rowcolor[gray]{0.95}Mean batch reward variance & $0.0596\;[0.0584,0.0609]$ & $0.0813\;[0.0803,0.0823]$ & $0.0726$--$0.0894$ \\
 Cumulative mean reward variance & $0.0534\;[0.0524,0.0543]$ & $0.0679\;[0.0673,0.0684]$ & $0.0639$--$0.0723$ \\
\rowcolor{envschedblue} Exploitation hit rate (\%) & $57.2\;[55.5,58.9]$ & $92.1\;[91.1,93.0]$ & $84.4$--$100.0$ \\
\bottomrule
\end{tabular*}
\begin{tabular*}{\textwidth}{@{\extracolsep{\fill}}ccccc@{}}
\toprule
\multicolumn{5}{l}{\textbf{(b) Persistence of the initial-batch effect}} \\
\midrule
\multirow{2}{*}{\textbf{Step}} & \multicolumn{2}{c}{\textbf{Mean batch reward variance}} & \multicolumn{2}{c}{\textbf{Exploitation hit rate (\%)}} \\
\cmidrule(lr){2-3}\cmidrule(lr){4-5}
& \textbf{Low initial group} & \textbf{High initial group} & \textbf{Low initial group} & \textbf{High initial group} \\
\midrule
0 & 0.0406 & 0.0540 & \multicolumn{2}{c}{N/A (initial batch)} \\
\rowcolor[gray]{0.95} 1 & 0.0574 & 0.0630 & 55.9 & 60.4 \\
4 & 0.0746 & 0.0768 & 84.6 & 83.4 \\
\rowcolor{envschedblue} 7 & 0.0810 & 0.0833 & 92.8 & 92.5 \\
\bottomrule
\end{tabular*}
\endgroup
\end{table*}

Across the 100 runs, the scheduler quickly moves toward high-signal tasks. Mean batch reward variance rises from 0.0596 at step 1 to 0.0813 at step 7, while the exploitation hit rate increases from 57.2\% to 92.1\%; even the 5th percentile step-7 branch reaches 84.4\%. The effect of the initial batch also decays rapidly. The high-initial-variance group starts 32.8\% above the low group, but the gap falls to 9.8\% after one step and 2.8\% by step 7. At the same point, exploitation hit rates are 92.8\% and 92.5\% for the low and high groups, respectively. These results show that random cold-start variation affects the first few updates but has little influence on the later task-selection quality.

\subsubsection{Neighborhood-Size Sensitivity}
We evaluate neighborhood-size sensitivity in complete 200-step training runs by independently halving or doubling either neighbor count around the default setting, $K_{\mathrm{within}}=6$ and $K_{\mathrm{cross}}=12$. As shown in Panel (b) of Table~\ref{tab:ablation-components-topk}, performance remains stable across these settings: the largest absolute deviations from the default are only 0.96, 0.81, and 1.33 percentage points on BFCL-v3, $\tau$-Bench, and ACEBench-Agent, respectively. All tested settings outperform random scheduling on all three benchmarks. This stability indicates that \envsched's gains are robust to the choice of neighborhood size within the tested range, with the default providing a reliable operating point without extensive hyperparameter tuning.

\section{Conclusion}
We presented \envsched, an environment-aware scheduler for reinforcement learning with executable agent environments. By treating environment--task pairs as the unit of data management, \envsched selects tasks with informative reward differentiation while maintaining broad environment coverage. Across Qwen3-1.7B, Qwen3-4B, Qwen3-8B, and Qwen3-30B-A3B, it consistently improves performance over random scheduling on BFCL-v3, $\tau$-Bench, and ACEBench-Agent under matched rollout budgets, with gains arising from the complementary effects of reward variance and environment diversity. More broadly, this work first establishes environment-level data management as a useful data-centric perspective for efficient agentic RL, paving the way for scaling increasingly capable agent models with better use of training rollout budgets.

\newpage

\bibliography{iclr2027_conference}\bibliographystyle{unsrtnat}

\newpage

\appendix

\section{Related Work}
\label{sec:related-work}

\subsection{Executable Environments for Agent Training}
Tool-use benchmarks provide executable settings for evaluating agents across function calling and multi-turn interaction. BFCL, $\tau$-Bench, and ACEBench assess complementary aspects of tool invocation, user interaction, and task completion~\citep{patil2025bfcl,yao2025taubench,chen2025acebench}. Recent environment-synthesis systems extend such interaction to diverse training domains. EnvFactory introduces 85 executable environments with 953 task instances for tool-using agents~\citep{xu2026envfactory}. EnvScaler~\citep{song2026envscaler} programmatically constructs environment skeletons, initial states, task scenarios, and terminal-state validation functions, producing 191 environments and approximately 7,000 scenarios. Agent World Model~\citep{wang2026awm} scales to 1,000 code-driven, database-backed environments with MCP tool interfaces and code-augmented reward verification. Agent-World~\citep{dong2026agentworld} combines environment--task discovery with multi-environment RL and a self-evolving arena that diagnoses capability gaps and synthesizes targeted tasks. Related work also studies embodied interaction and broad evaluation suites, including world-model interaction, embodied-agent benchmarking, and generative fluid-intelligence evaluation~\citep{chi2025wow,luo2026robobench,an2026genius}. These systems expand the supply of executable environments and verifiable tasks for agent training. \envsched addresses their use during training by organizing environment--task relationships and selecting informative, diverse tasks from the available pool.

\subsection{Task Informativeness Estimation and Online Selection}
Online task selection adapts the training distribution to the evolving policy by estimating task informativeness from reward variation, historical feedback, and task relationships. Recent technical reports on Agentic RL also treat asynchronous rollout, multi-task mixing, and environment scaling as central data-scheduling problems~\citep{zeng2026glm,mimo2026v26pro}. In group-relative RL, rollout reward differences identify informative tasks~\citep{shao2024deepseekmath}; DAPO~\citep{yu2025dapo} filters uniformly successful or unsuccessful groups and resamples prompts, while Pilot-Commit~\citep{kim2026pilotcommit} uses reward variation in a pilot sample to allocate later rollouts. For binary rewards, this informativeness is captured by variance $p(1-p)$. DOTS~\citep{sun2025dots} selects moderately difficult prompts using reference rollouts and an attention-based difficulty predictor, and Dynamics-Predictive Sampling (DPS)~\citep{mao2026dynamics} models prompt-solving dynamics from historical rollout outcomes to prioritize informative prompts before rollout generation. LEEPS~\citep{liang2026leeps} combines representation-space neighbors, historical outcomes, and explore--exploit portfolios to identify prompts likely to yield nonzero reward variance without extra rollouts, while graph-structured online difficulty estimation~\citep{liu2026graphdifficulty} propagates feedback over semantic and reasoning similarities through coupled latent difficulty states and a state-level Beta--Binomial model. Stable post-training recipes and controllable synthetic-data frameworks provide complementary evidence that data composition and training procedure jointly shape downstream capability~\citep{team2025nextstep,an2026concept}. Building on these foundations, \envsched incorporates executable environment structure through a dual-relation graph: within-environment edges capture shared tools, database fields, and verifier constraints, while cross-environment edges capture abstract execution and reward topology. These predictions guide task selection with explicit environment coverage, modeling both individual task informativeness and batch-level interaction patterns.

\section{Complete Historical Estimation}
\label{app:history-estimation}

This appendix gives the centering, clipping, and relation-specific neighbor weights used in Section~\ref{sec:task-informativeness}. For $y\in\{v,m,d\}$, let $b_s^y$ be the mean of $y$ over the tasks observed in training step $s$. The current global reference and the centered observation are

\begin{equation}
b_t^y=\frac{\sum_{s<t}w_{t,s}^{(h)}b_s^y}{\sum_{s<t}w_{t,s}^{(h)}},
\qquad
\widetilde y_{i,s}=\operatorname{clip}_{I_y}\!\left(y_{i,s}-b_s^y+b_t^y\right),
\label{eq:appendix-centering}
\end{equation}

where $I_v=[0,0.25]$ and $I_m=I_d=[0,1]$. If no step-level reference is available, the initialization reference is computed from the first observed batch. The task-history quantities in Eq.~\ref{eq:self-history} use these centered observations.

For the two edge types, $S^{\mathrm{in}}=S^{\mathrm{full}}$ and $S^{\mathrm{cross}}=S^{\mathrm{abs}}$. For relation type $q$, define the distance between tasks $i$ and $j$ as $\delta_{ij}^q=\operatorname{clip}_{[0,1]}(1-S_{ij}^q)$. The adaptive bandwidth and neighbor weight are

\begin{equation}
\sigma_{i,t}^{q,(h)}=\max\!\left\{\operatorname{median}_{j\in\mathcal{N}_t^q(i)}\delta_{ij}^q,0.05\right\},
\qquad
a_{ij,t}^{q,(h)}=\exp\!\left[-\frac{1}{2}\left(\frac{\delta_{ij}^q}{\sigma_{i,t}^{q,(h)}}\right)^2\right]
\sqrt{\operatorname{clip}_{[0.1,3]}(W_{j,t}^{(h)})}.
\label{eq:appendix-neighbor-weight}
\end{equation}

When a relation has no available observed neighbor, its estimate is set to the current global reference $b_t^y$. Percentile ranks are computed over the full task pool of $N$ environment--task pairs and use average ranks for ties:

\begin{equation}
\operatorname{pr}(z_i)=\frac{\operatorname{rank}_{\mathrm{avg}}(z_i)-0.5}{N}.
\label{eq:appendix-percentile}
\end{equation}

The empirical quantile function $Q_{\widehat v}$ is computed from the predicted variance values of all $N$ tasks and uses linear interpolation. The complete score in Eqs.~\ref{eq:informativeness-score}--\ref{eq:priority-score} is recomputed independently for the long and short half-lives, including centering, neighbor aggregation, percentile ranks, and quantile mapping.

\subsection{Observation Statistics and Priority Construction}

For the training implementation, each selected pair produces a complete group of $K$ rewards $r_{i,s,1},\ldots,r_{i,s,K}$. The overall reward mean, population variance, and mean-corrected dispersion are

\begin{equation}
\begin{aligned}
m_{i,s} &= \frac{1}{K}\sum_{k=1}^{K} r_{i,s,k},
&\qquad
v_{i,s} &= \frac{1}{K}\sum_{k=1}^{K}\left(r_{i,s,k}-m_{i,s}\right)^2, \\
d_{i,s} &= \operatorname{clip}_{[0,1]}\!\left(
\frac{v_{i,s}}{\max\{m_{i,s}(1-m_{i,s}),0.01\}}
\right).
\end{aligned}
\label{eq:task-statistics}
\end{equation}

These dispersion and clipping definitions use the training pool's reward scale $[0,1]$. Historical observations use the half-life weight

\begin{equation}
w_{t,s}^{(h)}=2^{-(t-s)/h}.
\label{eq:history-decay}
\end{equation}

For the centered observations in Eq.~\ref{eq:appendix-centering}, task histories are summarized as

\begin{equation}
W_{i,t}=\sum_{s\in\mathcal{H}_t(i)}w_{t,s}^{(h)},
\qquad
\bar y_{i,t}=\frac{\sum_{s\in\mathcal{H}_t(i)}w_{t,s}^{(h)}\widetilde y_{i,s}}{W_{i,t}},
\label{eq:self-history}
\end{equation}

Here $\mathcal{H}_t(i)$ contains earlier observation steps. Within a prediction pass using half-life $h$, $W_{i,t}$ and $\bar y_{i,t}$ denote the corresponding $W_{i,t}^{(h)}$ and $\bar y_{i,t}^{(h)}$. When no history is available, $W_{i,t}=0$ and $\bar y_{i,t}=b_t^y$. Predicted mean and dispersion reconstruct a second variance estimate, which is combined with the direct variance prediction:

\begin{equation}
f_{i,t}=\widehat d_{i,t}\widehat m_{i,t}(1-\widehat m_{i,t}),
\qquad
u_{i,t}^{(h)}=Q_{\widehat v}\!\left[0.65\operatorname{pr}(\widehat v_{i,t})+0.35\operatorname{pr}(f_{i,t})\right].
\label{eq:informativeness-score}
\end{equation}

All predicted quantities and the quantile function in this equation are computed independently for each half-life $h$. The final priority incorporates recent decline:

\begin{equation}
u_{i,t}=u_{i,t}^{(h_{\mathrm{long}})}-\eta_{\mathrm{decline}}\max\!\left\{u_{i,t}^{(h_{\mathrm{long}})}-u_{i,t}^{(h_{\mathrm{short}})},0\right\},
\qquad q_{i,t}=\operatorname{pr}(u_{i,t}).
\label{eq:priority-score}
\end{equation}

\section{Scheduling Rules and Cold Start}
\label{app:scheduling-rules}

The coverage phase first restricts the eligible candidates lexicographically by current-step coverage count, cumulative environment selections, and task observation count; it then applies Eq.~\ref{eq:batch-selection} using the informativeness score and batch similarity penalty. The hard constraints are parameterized by a per-round environment cap $C_1$, a rolling three-step environment cap $C_3$, and a consecutive-task cap $C_{\mathrm{task}}$, set to $C_1=4$, $C_3=11$, and $C_{\mathrm{task}}=2$ in our experiments.

The progress filter is part of candidate eligibility. It uses adjacent observations of the same task as training transitions. A ridge predictor estimates the later mean reward from the earlier mean reward, squared mean reward, variance, training step, observation count, and environment category. Each transition is predicted using only transitions from earlier completed training steps. The task's residual history is combined with environment and relation-neighbor references. A task with at least three scorable transitions and a non-positive estimated residual is removed from the exploitation candidates. It remains eligible for coverage and can return to exploitation after new observations change its estimate.

The identifier filter is also part of candidate eligibility. It examines high-reward trajectories for identifiers that cannot be obtained from the permitted context or for replacement objects that duplicate existing records. It first replays the saved fixed action sequence and checks that the replayed reward agrees with the stored reward within $10^{-4}$. It then removes the suspicious action and dependent edits and replays the sequence again. A task is filtered only when the removal lowers the reward. Replay uses the saved actions and does not sample a new model trajectory.

At cold start, EnvSched selects one representative task from each environment using the highest mean within-environment similarity. It fills the remaining slots by combining global novelty and within-environment novelty, with no more than two tasks from each environment in the first batch. This initialization uses static structure only.

\section{Static Representation and Runtime Configuration}
\label{app:implementation}

The task pool contains 2,550 task instances from 51 environments. The default batch contains $B=64$ environment--task pairs, with eight rollouts per pair. We use a fixed coverage fraction $\rho=0.5$, giving 32 coverage slots and 32 exploitation slots. The within-environment and cross-environment neighbor sizes are 6 and 12, respectively; the two relation estimates use fixed equal fusion weights of $0.5$ each, and the graph estimate and task-history prior use shrinkage strengths $\lambda_g=\lambda_s=2$. The long and short half-lives are $h_{\mathrm{long}}=8$ and $h_{\mathrm{short}}=2$ training steps, the recent-decline coefficient is $\eta_{\mathrm{decline}}=0.5$, the exposure-count scale is $c_0=16$, and the batch similarity penalty is $\lambda_{\mathrm{sim}}=0.08$. Environment caps are $C_1=4$, $C_3=11$, and $C_{\mathrm{task}}=2$. These hyperparameters are fixed across model sizes.

The six view weights are $(0.28,0.08,0.16,0.12,0.16,0.20)$. The six static views use word unigrams and bigrams, character $3$--$5$-grams, verifier-check text, verifier code structure, initial task context, and the complete operation description. Each view has its own TF--IDF encoder with minimum document frequency 1, a 35,000-feature cap, and sublinear term frequency. The structured operation description is extracted offline and normalized for the cross-environment representation. Numeric values are normalized into abstract categories so that cross-environment similarity reflects operation and condition types instead of exact identifiers.

\begingroup

\definecolor{EnvGraphBg}{HTML}{F8F9FA}
\definecolor{EnvGraphFrame}{HTML}{4A90D9}
\definecolor{EnvGraphText}{HTML}{2D3748}
\definecolor{EnvGraphBlue}{HTML}{3182CE}
\definecolor{EnvGraphGreen}{HTML}{38A169}
\definecolor{EnvGraphPurple}{HTML}{805AD5}
\definecolor{EnvGraphOrange}{HTML}{FAB387}
\definecolor{EnvGraphTeal}{HTML}{94E2D5}
\definecolor{EnvGraphCode}{HTML}{EDF2F7}

\newtcolorbox{EnvGraphPrompt}[2][robot]{
  enhanced,breakable,
  colback=EnvGraphBg,colframe=EnvGraphFrame,
  colbacktitle=EnvGraphFrame,coltitle=EnvGraphBg,
  fonttitle=\bfseries\sffamily,
  title={\faIcon{#1}\hspace{0.5em}#2},
  title after break={\faIcon{#1}\hspace{0.5em}#2},
  arc=2pt,boxrule=1.5pt,
  left=10pt,right=10pt,top=2pt,bottom=2pt,
  toptitle=0pt,bottomtitle=0pt,
  shadow={2pt}{-2pt}{0pt}{black!50},
  fontupper=\fontsize{9}{11}\selectfont\ttfamily\color{EnvGraphText},
  before upper={
    \raggedright
    \setlength{\parindent}{0pt}
    \setlength{\parskip}{0pt}
  },
  pad at break*=4pt,
}

\newcommand{\EnvGraphHeading}[3]{%
  \par\addvspace{6pt}\noindent
  {\color{#1}\faIcon{#2}\hspace{0.5em}\bfseries #3}%
  \par\nobreak\vspace{2pt}%
}

\newtcolorbox{EnvGraphInset}[1]{
  enhanced,colback=EnvGraphCode,colframe=EnvGraphCode,
  arc=3pt,boxrule=0pt,
  colupper=EnvGraphText,
  before upper={\raggedright},
  left=8pt,right=8pt,top=1pt,bottom=1pt,
  before skip=2pt,after skip=3pt,
}

\newcommand{\EnvGraphNumbered}[1]{%
  \par\noindent\hangindent=1.4em\hangafter=1
  #1\par\vspace{3pt}%
}

\newtcolorbox{EnvGraphEnumPanel}{
  enhanced,colback=EnvGraphCode,colframe=EnvGraphCode,
  arc=3pt,boxrule=0pt,
  colupper=EnvGraphText,
  before upper={\raggedright},
  left=7pt,right=7pt,top=5pt,bottom=5pt,
  before skip=2pt,after skip=3pt,
  fontupper=\fontsize{8.0}{9.5}\selectfont\ttfamily\color{EnvGraphText},
}

\newcommand{\EnvGraphEnumRow}[1]{%
  \par\noindent\hangindent=1.2em\hangafter=1
  #1\par\vspace{2.5pt}%
}

\newtcolorbox{EnvGraphCodePanel}[1]{
  enhanced,colback=EnvGraphCode,colframe=#1,
  arc=3pt,boxrule=1pt,
  colupper=EnvGraphText,
  before upper={\raggedright},
  left=8pt,right=8pt,top=1pt,bottom=1pt,
  before skip=3pt,after skip=3pt,
}

\newtcolorbox{EnvGraphInputPanel}{
  enhanced,colback=white,colframe=EnvGraphTeal,
  colupper=EnvGraphText,
  arc=2pt,boxrule=0.5pt,
  left=10pt,right=10pt,top=8pt,bottom=8pt,
  before upper={\raggedright},
  before skip=3pt,after skip=3pt,
}

\newcommand{\EnvGraphRule}[1]{%
  \par\noindent\hangindent=1.2em\hangafter=1
  #1\par\vspace{3.5pt}%
}

\newenvironment{EnvGraphRulesBlock}
  {\par\noindent\begin{minipage}{\linewidth}\raggedright}
  {\end{minipage}\par}

\lstdefinestyle{EnvGraphJSON}{
  basicstyle=\fontfamily{cmtt}\fontsize{7.8}{9.4}\selectfont
    \color{EnvGraphText},
  columns=fullflexible,
  keepspaces=true,
  showstringspaces=false,
  breaklines=true,
  breakatwhitespace=false,
  breakindent=1em,
  aboveskip=0pt,
  belowskip=0pt,
  upquote=true,
  morestring=[b]",
  stringstyle=\color{EnvGraphText},
  literate={},
}

\lstnewenvironment{EnvGraphJSON}
  {\lstset{style=EnvGraphJSON}}
  {}

\section{Cross-Environment Graph Construction Prompt}
\label{app:topology-extraction}

This appendix provides the complete prompt used to construct cross-environment edges in the dual-relation graph (Section~\ref{sec:dual-relation-graph}). Within-environment edges use TF--IDF over concrete features like tool names and verifier code. Cross-environment edges require an abstract representation that removes domain-specific identifiers.

\subsection{What This Prompt Extracts}

Given a task specification (instruction, initial state, verifier checks, available tools), the prompt outputs a structured description of what operations the task requires and how they depend on each other. After removing domain-specific identifiers, this structure becomes the cross-environment similarity $S^{\mathrm{abs}}$ in Eq.~\ref{eq:task-similarities}.

The prompt outputs five components:

\begin{enumerate}
\item \textbf{Operations}: The atomic actions required to complete the task. Each operation is labeled with a type that describes its effect:
\begin{itemize}
\item \texttt{lookup}: retrieve data from the environment (e.g., ``find user record by username'')
\item \texttt{update}: modify existing records (e.g., ``change user status field to active'')
\item \texttt{create}: add new entities to the environment (e.g., ``create new reservation'')
\item \texttt{delete}: remove records
\item \texttt{validate}: verify constraints without changing state
\item \texttt{authorize}: check permissions
\end{itemize}
Each operation lists which other operations must complete before it can start. For example, ``update\_user\_status depends on lookup\_user because the user ID from lookup is needed as input to update.''

\item \textbf{Gates}: Logical conditions that must be satisfied before operations can proceed. If a gate cannot be satisfied, all operations depending on it are blocked. Three examples illustrate the concept:
\begin{itemize}
\item \emph{Identifier resolution gate}: The task says ``update status for user Alice'', but the update operation requires a numeric user ID, not a name. A gate captures this requirement: the user ID must first be obtained by looking up ``Alice'' in the user table.
\item \emph{State precondition gate}: The task says ``assign patient to bed 301'', but the assignment operation requires that bed 301 is currently unoccupied. A gate captures this requirement: the bed's occupancy status must be checked before assignment can proceed.
\item \emph{Authorization gate}: The task says ``delete the billing record'', but the delete operation requires admin privileges. A gate captures this requirement: the agent must have the required permission, which may not be available in the current environment.
\end{itemize}
Each gate also records its \textbf{provenance} (where the required information comes from):
\begin{itemize}
\item \texttt{task\_literal}: the task instruction directly supplies the value (e.g., ``update user ID 42'')
\item \texttt{unique\_lookup}: a tool call deterministically finds it (e.g., username ``alice'' matches exactly one user)
\item \texttt{broad\_lookup}: multiple records match; the agent must choose (e.g., name ``John Smith'' matches 3 users)
\item \texttt{prior\_tool\_output}: an earlier operation produces it (e.g., \texttt{create\_user} returns the new user ID)
\item \texttt{agent\_generated}: the agent must create the value (e.g., generate a password)
\item \texttt{unavailable}: no available tool can obtain it
\end{itemize}

\item \textbf{Dependencies}: Directed edges stating which operations must finish before others can start. For example, if operation B needs the output of operation A as input, then B depends on A. This is recorded as an edge from A to B in the operation graph.

\item \textbf{Reward groups}: A partition of the verifier checks (the success conditions for the task). Verifier checks that are controlled by the same operation or gate belong to the same group, because they will succeed or fail together. Three examples:
\begin{itemize}
\item Suppose the verifier has three checks: ``user status is active'', ``user last\_login timestamp is updated'', and ``user notification\_sent flag is true''. If all three fields are written by the same \texttt{update\_user} operation, the three checks belong to one group with relation type \texttt{shared\_operation}.
\item Suppose two verifier checks both examine properties of a user record, and obtaining that user record requires satisfying an identifier-resolution gate. If the gate fails, both checks fail. They belong to one group with relation type \texttt{shared\_gate}.
\item Suppose one verifier check examines a field that already has the correct value in the initial state, and no operation in the task will modify it. This check is already satisfied before the agent does anything. It belongs to a group with relation type \texttt{initially\_satisfied}.
\end{itemize}

\item \textbf{Conflicts}: Pairs of operations that have ordering constraints or incompatibilities beyond simple data dependencies. Three conflict types:
\begin{itemize}
\item \texttt{order}: Operation A must execute before operation B, even if B does not directly consume A's output. For example, ``verify eligibility must precede commit transaction'' ensures that invalid transactions are never committed, even though the commit operation does not use the verification result as a data input.
\item \texttt{mutual\_exclusion}: Operations A and B cannot both execute in the same task. For example, ``approve request'' and ``reject request'' are mutually exclusive: completing the task requires executing one or the other, but not both.
\item \texttt{destructive}: Operation A removes the preconditions needed by operation B. For example, \texttt{delete\_user} makes any subsequent \texttt{update\_user} impossible because the user record no longer exists.
\end{itemize}
\end{enumerate}

After extraction, we apply a deterministic abstraction procedure to the output:
\begin{enumerate}
\item Replace each tool name with its operation type (e.g., \texttt{get\_user\_by\_name} $\to$ \texttt{lookup}, \texttt{update\_user\_status} $\to$ \texttt{update})
\item Remove all entity identifiers (user IDs, record keys, table names)
\item Remove all database field names (e.g., \texttt{status}, \texttt{assignment\_id})
\item Replace literal values with type placeholders (e.g., ``active'' $\to$ \texttt{<string>}, 42 $\to$ \texttt{<int>})
\item Retain the dependency structure, gate types, and group relations
\end{enumerate}
This produces an abstract structure description. For example, a task ``update user Alice's status to active'' with concrete extraction \texttt{[\allowbreak{}lookup(\allowbreak{}get\_\allowbreak{}user\_\allowbreak{}by\_\allowbreak{}name,\allowbreak{} input=\allowbreak{}Alice),\allowbreak{} update(\allowbreak{}set\_\allowbreak{}user\_\allowbreak{}status,\allowbreak{} target=\allowbreak{}user\_\allowbreak{}id,\allowbreak{} value=\allowbreak{}active)]} becomes abstract structure \texttt{[\allowbreak{}lookup(\allowbreak{}type=\allowbreak{}entity\_\allowbreak{}by\_\allowbreak{}name),\allowbreak{} update(\allowbreak{}type=\allowbreak{}entity\_\allowbreak{}field)]} after applying these replacements. The abstract string is TF--IDF encoded to produce $g_i$ in Eq.~\ref{eq:task-similarities}, which computes cross-environment similarity $S_{ij}^{\mathrm{abs}}$ for graph edges $\mathcal{E}^{\mathrm{cross}}$.

\clearpage
\subsection{System Prompt}

The prompt constrains the model to use fixed vocabularies for operation types, gate types, and group relations. JSON structure is enforced by \texttt{response\_format=\{"type":"json\_object"\}} in the API call. 

\begin{EnvGraphPrompt}{System Prompt}
You extract factual static topology from tool-agent tasks. You receive no rollout,
trajectory, action trace, model output, reward observation, success rate, or variance.
Do not predict any of them and do not assign difficulty, probability, capability, or
high/low labels.

\par
\EnvGraphHeading{EnvGraphPurple}{bullseye}{Build two connected structures}
\begin{EnvGraphInset}{EnvGraphPurple}
 from the supplied task, initial state, exact verifier
checks, tool catalog, and selected real tool implementations:

\par\smallskip
\EnvGraphNumbered{1. execution topology: the minimum correct operations, their dependencies, and factual
   prerequisite gates such as identifier provenance, state preconditions, ordering,
   authorization, exact arguments, and missing tool affordances;
}
\EnvGraphNumbered{2. reward topology: a partition of verifier check IDs into groups that are controlled
   by the same operation or prerequisite gate. Check weights are not yours to estimate;
   downstream code computes them from the exact check IDs.
}
\end{EnvGraphInset}

\EnvGraphHeading{EnvGraphBlue}{key}{A gate is a prerequisite}
 whose resolution can jointly enable or block one or more
operations/checks. Do not create a gate for every ordinary literal. Distinguish a
direct task literal, unique lookup, broad lookup, prior tool output, agent-generated
value, and genuinely unavailable value. Mark reachability unsupported only when the
supplied tools/implementation establish that no path exists; otherwise use unknown.
Describe topology, not likely model behavior.

\par
\EnvGraphHeading{EnvGraphTeal}{database}{The initial\_state\_summary}
\begin{EnvGraphInset}{EnvGraphOrange}
 is evidence available to this static analyzer, not
information shown to the agent. Never mark an identifier direct merely because it
appears in that summary. From the agent\textquotesingle{}s execution perspective, an identifier is
task\_literal only when the task supplies the execution-ready value; otherwise classify
the actual tool path as unique\_lookup, broad\_lookup, prior\_tool\_output, inferred, or
unavailable. Use the initial state only to verify that records, matches, and
preconditions exist and to determine whether a supplied lookup can resolve them.
\end{EnvGraphInset}

\EnvGraphHeading{EnvGraphBlue}{list-ol}{Use these enums exactly:}
\begin{EnvGraphEnumPanel}
\EnvGraphEnumRow{- operation.family: [\textquotesingle{}aggregate\textquotesingle{}, \textquotesingle{}authorize\textquotesingle{}, \textquotesingle{}create\textquotesingle{}, \textquotesingle{}delete\textquotesingle{}, \textquotesingle{}link\textquotesingle{}, \textquotesingle{}lookup\textquotesingle{},
  \textquotesingle{}other\textquotesingle{}, \textquotesingle{}report\textquotesingle{}, \textquotesingle{}schedule\textquotesingle{}, \textquotesingle{}transition\textquotesingle{}, \textquotesingle{}update\textquotesingle{}, \textquotesingle{}validate\textquotesingle{}]
}
\EnvGraphEnumRow{- gate.type: [\textquotesingle{}actor\_authorization\textquotesingle{}, \textquotesingle{}argument\_exactness\textquotesingle{}, \textquotesingle{}capacity\_or\_eligibility\textquotesingle{},
  \textquotesingle{}cross\_entity\_join\textquotesingle{}, \textquotesingle{}generated\_identifier\textquotesingle{}, \textquotesingle{}identifier\_resolution\textquotesingle{}, \textquotesingle{}ordering\textquotesingle{},
  \textquotesingle{}other\textquotesingle{}, \textquotesingle{}state\_precondition\textquotesingle{}, \textquotesingle{}temporal\_conversion\textquotesingle{}, \textquotesingle{}termination\_horizon\textquotesingle{},
  \textquotesingle{}tool\_affordance\textquotesingle{}]
}
\EnvGraphEnumRow{- gate.provenance: [\textquotesingle{}agent\_generated\textquotesingle{}, \textquotesingle{}broad\_lookup\textquotesingle{}, \textquotesingle{}inferred\textquotesingle{}, \textquotesingle{}none\textquotesingle{},
  \textquotesingle{}prior\_tool\_output\textquotesingle{}, \textquotesingle{}task\_literal\textquotesingle{}, \textquotesingle{}unavailable\textquotesingle{}, \textquotesingle{}unique\_lookup\textquotesingle{}, \textquotesingle{}unknown\textquotesingle{},
  \textquotesingle{}wall\_clock\textquotesingle{}]
}
\EnvGraphEnumRow{- gate.reachability: [\textquotesingle{}ambiguous\textquotesingle{}, \textquotesingle{}direct\textquotesingle{}, \textquotesingle{}indirect\textquotesingle{}, \textquotesingle{}unknown\textquotesingle{}, \textquotesingle{}unsupported\textquotesingle{}]
}
\EnvGraphEnumRow{- gate.observability: [\textquotesingle{}explicit\_result\textquotesingle{}, \textquotesingle{}silent\textquotesingle{}, \textquotesingle{}state\_query\textquotesingle{}, \textquotesingle{}unknown\textquotesingle{}]
}
\EnvGraphEnumRow{- gate.recoverability: [\textquotesingle{}irreversible\textquotesingle{}, \textquotesingle{}retryable\textquotesingle{}, \textquotesingle{}unknown\textquotesingle{}]
}
\EnvGraphEnumRow{- reward\_group.relation: [\textquotesingle{}fixed\_unreachable\textquotesingle{}, \textquotesingle{}independent\textquotesingle{}, \textquotesingle{}initially\_satisfied\textquotesingle{},
  \textquotesingle{}mixed\textquotesingle{}, \textquotesingle{}shared\_gate\textquotesingle{}, \textquotesingle{}shared\_operation\textquotesingle{}]
}
\EnvGraphEnumRow{- conflict.kind: [\textquotesingle{}destructive\textquotesingle{}, \textquotesingle{}mutual\_exclusion\textquotesingle{}, \textquotesingle{}order\textquotesingle{}, \textquotesingle{}preservation\textquotesingle{}]
}
\end{EnvGraphEnumPanel}

\EnvGraphHeading{EnvGraphGreen}{file-export}{Return JSON with key \textquotedbl{}tasks\textquotedbl{}.}
 Every task item must have exactly this structure:

\begin{EnvGraphCodePanel}{EnvGraphGreen}
\begin{EnvGraphJSON}
{
  "task_id": "...",
  "operations": [
    {"id":"o0","family":"lookup","description":"...","tool_candidates":["..."],
      "depends_on_operations":[],"gate_ids":["g0"],"affected_checks":["c0"]}
  ],
  "gates": [
    {"id":"g0","type":"identifier_resolution","description":"...",
      "provenance":"unique_lookup","reachability":"indirect",
      "observability":"explicit_result","recoverability":"retryable",
      "depends_on_gates":[],"affected_operations":["o0"],"affected_checks":["c0"]}
  ],
  "reward_groups": [
    {"id":"r0","check_ids":["c0"],"shared_gate_ids":["g0"],
      "producer_operations":["o0"],"relation":"shared_gate"}
  ],
  "conflicts": [{"left_operation":"o0","right_operation":"o1","kind":"order"}],
  "independent_branches": [
    {"id":"b0","operation_ids":["o0"],"check_ids":["c0"]}
  ],
  "uncertainties": ["short factual ambiguity in the supplied static evidence"]
}
\end{EnvGraphJSON}
\end{EnvGraphCodePanel}

\begin{EnvGraphRulesBlock}
\EnvGraphHeading{EnvGraphOrange}{clipboard-check}{Rules:}

\EnvGraphRule{- Use only supplied check IDs and tool names.
}
\EnvGraphRule{- reward\_groups must partition every verifier check exactly once. Group checks only
  when they share a concrete gate/producer; otherwise use singleton or independent
  groups.
}
\EnvGraphRule{- A check already true in the untouched initial state belongs in an initially\_satisfied
  group unless later required operations can invalidate it.
}
\EnvGraphRule{- Operations describe the shortest correct state-changing/query plan, not speculative
  retries.
}
\EnvGraphRule{- IDs must be local sequential identifiers o0..., g0..., r0..., b0....
}
\EnvGraphRule{- Keep descriptions concise and case-specific. At most 30 operations, 30 gates, 30
  reward groups, 20 conflicts, 20 branches, and 5 uncertainties per task.
}
\EnvGraphRule{- Return JSON only.
}
\end{EnvGraphRulesBlock}
\end{EnvGraphPrompt}

\subsection{User Message Format}

The user message is structured JSON with no additional natural language. Tasks from the same environment share a \texttt{shared\_environment} object:

\begin{EnvGraphPrompt}[terminal]{User Message Format}
\begin{EnvGraphInputPanel}
\begin{EnvGraphJSON}
{
  "shared_environment": {
    "environment": "<environment class name>",
    "summary": "<prose summary>",
    "state_space": "<state definition>",
    "constraints": "<rules>",
    "tool_catalog": [
      {"name": "...", "description": "...", "required": [...],
       "arguments": [...], "signature": "..."}
    ]
  },
  "tasks": [
    {
      "task_id": "<identifier>",
      "task": "<instruction>",
      "initial_state_summary": {<generated by summarize_initial_state>},
      "verifier_checks": [
        {"check_id": "c0", "requirement": "...", "initially_true": <bool>,
         "verifier_code": "..."}
      ],
      "relevant_tool_implementations": [
        {"name": "...", "signature": "...", "documentation": "...",
         "implementation": "..."}
      ]
    }
  ]
}
\end{EnvGraphJSON}
\end{EnvGraphInputPanel}
\end{EnvGraphPrompt}

\endgroup

\section{Qualitative Case Studies}
\label{app:case-studies}

This appendix presents three cases from the Qwen3-4B run, one per subsection. Each case is written as the scheduler saw it: the task, the evidence available before the decision, the prediction that evidence produced, and the eight rollouts that followed. All three are tasks that had never been selected before, so nothing was known about the task itself and every prediction came from other environment--task pairs. Neighbor tables list the representative neighbors each relation retrieved, their similarity, the number of prior rollouts, and the variance each contributes. The deployed estimates use the full neighbor sets ($K_{\mathrm{within}}=6$, $K_{\mathrm{cross}}=12$).

\subsection{Within-Environment Edges: \texttt{env\_141-task\_27} at Step 28}
\label{app:case-within}

Two tasks in the same environment share tools, database fields, and verifier conditions. When one of them has been rolled out, its outcome is therefore evidence about the other: a mistake in the shared edit sequence costs both of them, and the checks that depend on the same steps fail in both.

\textbf{The task.} InsurancePolicyManagementSystem. Acting as Carolyn Ho, amend two policies: set the active period of policy OXL-44032-F, raise its hospitalization limit to 30{,}000 and lower its deductible to 400, add an Outpatient item and a Dental item while removing the Dental exclusion and keeping the Elective exclusion; then transfer policy ENL-74522-K to another account, set its active period, and add a Psychiatric exclusion. Sixteen verifier checks. The task has the characteristic policy-editing shape: an authorization check on the acting account, lookups that resolve policy numbers and coverage-item identifiers, paired deletions and creations of coverage items, and field updates whose order matters because the tool rejects a duplicate coverage type.

\textbf{What the scheduler knew.} The task itself had never been selected. Its environment was the most heavily sampled in the pool: 56 prior groups spanning 40 distinct tasks, with mean reward variance 0.055 against a pool-wide reference of 0.026 at this step. All six within-environment neighbors had been rolled out.

\begin{center}
\begin{tabular*}{\linewidth}{@{\extracolsep{\fill}}lccccc@{}}
\toprule
Within-environment neighbor & Similarity & Observations & Last step & Variance & Weight \\
\midrule
\texttt{task\_5}  & 0.554 & 2 & 26 & 0.0146 & 0.125 \\
\texttt{task\_38} & 0.544 & 1 & 19 & 0.0077 & 0.086 \\
\texttt{task\_18} & 0.525 & 1 & 25 & 0.0414 & 0.107 \\
\texttt{task\_19} & 0.508 & 1 & 11 & 0.0000 & 0.056 \\
\texttt{task\_17} & 0.506 & 2 & 21 & 0.1863 & 0.118 \\
\texttt{task\_46} & 0.506 & 1 & 27 & 0.1084 & 0.112 \\
\bottomrule
\end{tabular*}
\end{center}

The neighbors are not merely similar-sounding requests. They call the same functions on the same tables as the target, and all of them are subject to the same ordering constraint that the coverage-item tool rejects a duplicate coverage type.

\begin{center}
\begin{tabularx}{\linewidth}{@{}l>{\raggedright\arraybackslash}Xc@{}}
\toprule
Neighbor & Tools shared with the target & Checks \\
\midrule
\texttt{task\_17} (var.\ 0.186) & \texttt{add\_coverage\_item}, \texttt{remove\_exclusion}, \texttt{update\_coverage\_limit\_or\_deductible}, \texttt{update\_policy\_active\_dates} & 12 \\
\texttt{task\_46} (var.\ 0.108) & \texttt{add\_coverage\_item}, \texttt{remove\_exclusion}, \texttt{update\_coverage\_limit\_or\_deductible}, \texttt{update\_policy\_active\_dates} & 8 \\
\texttt{task\_5} (var.\ 0.015) & \texttt{add\_coverage\_item}, \texttt{remove\_exclusion}, \texttt{update\_coverage\_limit\_or\_deductible}, \texttt{update\_policy\_active\_dates} & 9 \\
\bottomrule
\end{tabularx}
\end{center}

\textbf{The prediction.} The within-environment relation averages these six neighbors to $0.05$. The cross-environment relation, which retrieves tasks from other environments by shape, averages to $0.02$, and the global reference is $0.03$. The two are fused with equal weight (Eq.~\ref{eq:history-fusion}) to $0.04$.

\textbf{The selection.} Exploitation phase, batch rank 38, predicted percentile $q = 0.9747$, predicted mean reward 0.579 and predicted dispersion 0.194.

\textbf{What the rollouts showed.} $(0.1875, 0.1875, 0.1875, 0.1875, 0, 0.750, 0.750, 0.625)$; mean 0.359, variance 0.078, against a batch mean realized variance of 0.036 at this step. Seven of the eight trajectories form two clear behavioral groups, with one zero-reward exception, and the difference between them is not a matter of degree but of whether the trajectory attempts the edit sequence at all. Four trajectories call only \texttt{is\_authorized\_user\_for\_policy}, one or two calls each, and then stop. They pass three of the sixteen checks: the two authorization checks, and c10, which asks whether the Elective exclusion has been left in place --- a check that a trajectory passes by doing nothing. The other thirteen checks fail. The three high trajectories carry out the full sequence of lookups, coverage-item insertions, exclusion removals, and the policy transfer, reaching 12, 12, and 10 of 16 checks, with what remains failing on the second policy's start date and its transfer. The single zero trajectory makes six calls, including \texttt{update\_policy\_active\_dates} and \texttt{update\_coverage\_limit\_or\_deductible}, and then ends before reaching the verifier. 

\subsection{Cross-Environment Edges: \texttt{env\_190-task\_8} at Step 40}
\label{app:case-cross}

Two tasks in different domains can ask for the same kind of work. Reconcile a patient's allergy list, clean up a member's mood history, and approve two expense claims have nothing in common in what they are about, but all three ask the agent to identify a record from the request, establish that it is allowed to touch it, and then apply a short list of edits to that record and nothing else. When two tasks are put together that way, they also tend to be hard or easy in the same places, which is what makes one informative about the other despite the different subject matter.

\textbf{The task.} ElectronicHealthRecordSystem. As Dr. Marta Fong, reconcile the allergy list of a patient: confirm authorization, locate the patient by demographics, update an existing Sulfa Drugs allergy, remove a Latex entry, add Adhesive Tape and Penicillin entries with exact reaction types and confirmation statuses, and leave no duplicates. Nothing can be edited until the patient has been located and the doctor's permission has been confirmed, so a single lookup near the start decides whether the rest of the task is reachable at all.

\textbf{What the scheduler knew.} No own history. The environment had been sampled 48 times over 26 distinct tasks, but the four nearest same-environment tasks had observed variances of 0.0000, 0.0123, 0.0032, and 0.0000: the earlier tasks here, which asked for the same kind of working over a patient's records, had all come out the same way. The four nearest tasks outside the environment sit in a mood tracker, an emergency-response app, an expense system, and a video-conferencing scheduler. They involve different records, different tools, and different verifiers, and none of them is about allergies.

\begin{center}
\begin{tabularx}{\linewidth}{@{}l>{\raggedright\arraybackslash}Xr@{}}
\toprule
Neighbor & What it asks the agent to do & Variance \\
\midrule
\texttt{env\_177-task\_21} & delete one mood entry, edit a second, add a third & 0.1160 \\
\texttt{env\_164-task\_11} & cancel one session, reschedule another, revise its participants & 0.0830 \\
\texttt{env\_175-task\_35} & report, resolve, retract, and update incidents & 0.0381 \\
\texttt{env\_161-task\_32} & assign an approver, verify a document, advance a claim & 0.0256 \\
\bottomrule
\end{tabularx}
\end{center}

What these tasks share, despite the different subject matter, is the work they ask for: find the record the request names, establish that the agent is allowed to touch it, then change that record in a few specific ways. The target reads the acting user's rights, locates one patient, and edits that patient's allergy entries, changing one, removing another, and adding two. \texttt{env\_177-task\_21} does the same thing to a member's mood history: one entry deleted, one edited, one added. \texttt{env\_164-task\_11} works on sessions in the same manner, cancelling one and rescheduling another while revising its participant list. \texttt{env\_161-task\_32} advances a claim by assigning an approver, verifying a document, and moving its status forward, and \texttt{env\_175-task\_35} reports, resolves, retracts, and updates incidents.

The consequence of this arrangement shows up in the target's own rollouts. Because the four allergy edits all belong to the same patient and every tool takes that patient identifier, a rollout that cannot resolve the patient cannot make any of the four: five of the eight never do and score zero, while the three that succeed are separated only by which of the edits they finish. The neighbor tasks in the table are put together the same way, and the variance beside each is the variance those tasks actually produced.

This is why these tasks can inform one another even though none of them is about allergies. What is compared is not what the records mean or what the tools are called, but the shape of the work: which record has to be found before anything can happen, what is done to it, and how the task is arranged around it.

\textbf{The prediction.} The cross-environment relation averages these neighbors to $0.04$. The within-environment relation averages to $0.02$ from neighbors whose observed variances are 0.0000, 0.0123, 0.0032, and 0.0000. Those near-zero variances are not missing information: they report that the earlier ElectronicHealthRecordSystem tasks were solved the same way every time. What they cannot report is whether a different patient's allergy list, with different entries and different conflicts, will be as easy, because every task they summarize is drawn from this one environment. The cross-environment neighbors cover the tasks in other environments that are put together the same way, and their variances --- 0.0830, 0.0381, 0.0256, 0.1160 --- say how much such a task tends to separate the policy. The two are fused with equal weight to $0.03$ (Eq.~\ref{eq:history-fusion}).

\textbf{The selection.} Exploitation phase, batch rank 33, $q = 0.9786$, predicted mean reward 0.562 and predicted dispersion 0.136.

\textbf{What the rollouts showed.} $(0.2143, 0.2143, 0.7857, 0, 0, 0, 0, 0)$; mean 0.152, variance 0.066, against a batch mean realized variance of 0.038 at this step. All five zero trajectories end before the verifier after hitting the 40-turn limit, and each one dies in the same place: alternating \texttt{get\_patient\_by\_demographics} and \texttt{verify\_user\_authorization} across 40, 40, 40, 40, and 33 calls without ever obtaining a usable patient identifier or authorization result, so no downstream edit is attempted. The single 0.7857 trajectory clears the identity and authorization steps, calls \texttt{get\_all\_patients} and \texttt{list\_patient\_allergies}, and completes the allergy edits for 11 of 14 checks. The two 0.2143 trajectories reach the verifier having satisfied only the three identity and authorization checks. The failure mode is the one the cross-environment neighbors pointed to: in all five, a single unresolved lookup blocks every downstream check at once.

\subsection{Both Relations Contributing: \texttt{env\_156-task\_5} at Step 4}
\label{app:case-both}

The two relations are used together rather than as alternatives. On a task where both find observed neighbors, the fused estimate is what determines where the task lands in the ranking.

\textbf{The task.} BloodDonationAppointmentSystem, early in training. Reschedule two donors' appointments at their existing clinics, release the vacated slots and block the new ones, release a slot from a cancelled appointment, confirm a fourth donor's eligibility, and create his appointment. Twelve checks. The shape is the many-lookup one typical of early cold-start batches: the agent must resolve facility names and time slots before it can change any state, so a small number of edits sits behind a long chain of lookups.

\textbf{What the scheduler knew.} No own history, and the environment itself had only five prior groups over four tasks, so its own tasks offer little. Both relations nonetheless retrieve recorded neighbors.

\begin{center}
\setlength{\tabcolsep}{4pt}
\begin{tabular*}{\linewidth}{@{\extracolsep{\fill}}lccccc@{}}
\toprule
Neighbor & Relation & Similarity & Observations & Variance & Weight \\
\midrule
\texttt{env\_156-task\_39} & within & 0.450 & 1 & 0.1073 & 0.131 \\
\texttt{env\_156-task\_10} & within & 0.372 & 2 & 0.1093 & 0.184 \\
\texttt{env\_143-task\_47} & cross & 0.630 & 1 & 0.2013 & 0.065 \\
\texttt{env\_145-task\_34} & cross& 0.614 & 1 & 0.1552 & 0.055 \\
\texttt{env\_157-task\_8} & cross& 0.619 & 1 & 0.1036 & 0.063 \\
\bottomrule
\end{tabular*}
\end{center}

The two within-environment neighbors share this environment's clinic and time-slot tools. The cross-environment neighbors come from hospital patient management, a review platform, and a licensing board, and share the many-lookup pattern in which a broad lookup by name decides whether anything downstream can proceed.

\textbf{The prediction.} The within-environment relation gives $0.07$ and the cross-environment relation $0.10$, fusing to $0.09$ against a global reference of $0.06$ (Eq.~\ref{eq:history-fusion}).

\textbf{The selection.} Exploitation phase, batch rank 40, $q = 0.9861$, predicted mean reward 0.505 and predicted dispersion 0.387.

\textbf{What the rollouts showed.} $(1, 1, 0.9167, 0.9167, 0.0833, 0.0833, 0.0833, 0)$; mean 0.510, variance 0.202, twice the batch mean realized variance of 0.091 at this step. The two perfect trajectories execute the full schedule in 24 and 26 calls and pass all twelve checks. The two 0.9167 trajectories differ from them only on c9, the creation of the fourth donor's appointment. The three 0.0833 trajectories carry out the rescheduling blocks and then fail the downstream checks. The zero trajectory spends its 40 turns alternating \texttt{list\_locations} and \texttt{get\_location\_by\_name} without ever resolving a facility name --- the same lookup bottleneck the cross-environment neighbors point to.

\end{document}